\documentclass[runningheads]{llncs}

\usepackage[T1]{fontenc}
\usepackage{graphicx}

\usepackage{soul}
\usepackage{url}
\usepackage[hidelinks]{hyperref}
\usepackage[utf8]{inputenc}
\usepackage{graphicx}
\usepackage{amsmath}
\usepackage{booktabs}

\usepackage[]{cleveref}
\usepackage{amssymb}
\usepackage[caption=false]{subfig}
\usepackage{multirow} 
\usepackage{stmaryrd}
\usepackage{pifont}%
\newcommand{\cmark}{\ding{51}}%
\usepackage{xcolor} %
\usepackage{wrapfig}

\newcommand\ours{DiPE-Linear}
\newcommand\invfft{\mathcal{F}^{-1}}
\newcommand\fft{\mathcal{F}}

\newcommand{\best}[1]{\textbf{\textcolor{red}{#1}}}

\newcommand{\second}[1]{\underline{\textcolor{blue}{#1}}}

\usepackage{color}

\usepackage{marvosym}
\newcommand\blfootnote[1]{%
  \begingroup
  \renewcommand\thefootnote{}\footnote{#1}%
  \addtocounter{footnote}{-1}%
  \endgroup
}

\begin{document}
\title{Disentangled Parameter-Efficient Linear Model for Long-Term Time Series Forecasting}
\titlerunning{Disentangled Parameter-Efficient Linear Model for LTSF}

\author{
Yuang Zhao\inst{1,2}
\and
Tianyu Li\inst{2}
\and
Jiadong Chen\inst{1}
\and
Shenrong Ye\inst{2}
\and \\
Fuxin Jiang\textsuperscript{(\Letter),}\inst{3}
\and
Xiaofeng Gao\textsuperscript{(\Letter),}\inst{1}
}
\authorrunning{Y. Author et al.}
\institute{
Shanghai Key Laboratory of Scalable Computing and Systems, \\School of Computer Science, Shanghai Jiao Tong University, Shanghai, China\\
\email{\{zhaoyuang, chenjiadong998\}@sjtu.edu.cn, gao-xf@cs.sjtu.edu.cn}
\and
SJTU Paris Elite Institute of Technology,\\ Shanghai Jiao Tong University, Shanghai, China\\
\email{\{hugo\_li, olivier.9928\}@sjtu.edu.cn}
\and
ByteDance Inc., Beijing, China\\
\email{jiangfuxin@bytedance.com}
}

\maketitle

\blfootnote{(\Letter)\ Corresponding authors. 
This work was supported by the National Key R\&D Program of China [2024YFF0617700], the National Natural Science Foundation of China [U23A20309, 62272302, 62372296], and ByteDance Research Project [CT20241217115379].}

\begin{abstract}
Long-term Time Series Forecasting (LTSF) is crucial across various domains, but complex deep models like Transformers are often prone to overfitting on extended sequences. Linear Fully Connected (FC) models have emerged as a powerful alternative, achieving competitive results with fewer parameters. However, their reliance on a single, {monolithic} weight matrix leads to quadratic parameter redundancy and an {entanglement} of temporal and frequential properties. To address this, we propose \textbf{\ours}, a novel model that {disentangles} this monolithic mapping into a sequence of specialized, parameter-efficient modules. \ours~features three core components: {Static Frequential Attention} to prioritize critical frequencies, {Static Time Attention} to focus on key time steps, and {Independent Frequential Mapping} to independently process frequency components. A {Low-rank Weight Sharing} policy further enhances efficiency for multivariate data. This disentangled architecture collectively reduces parameter complexity from quadratic to linear and computational complexity to log-linear. Experiments on real-world datasets show that \ours~delivers state-of-the-art performance with significantly fewer parameters, establishing a new and highly efficient baseline for LTSF. Our code is available at \url{https://github.com/wintertee/DiPE-Linear/}

\end{abstract}
\section{Introduction}

Long-term Time Series Forecasting (LTSF) plays a crucial role in a wide array of fields, such as weather forecasting \cite{angryk2020multivariate} and industrial manufacturing \cite{zhou2021attention}, making it a prominent focus of research in real-world applications. However, LTSF presents significant challenges due to the need for processing extended look-back window and forecasting horizon, which increases both the complexity of the forecasting task and the dimensionality of the model.
Recent advancements in deep learning, such as 
Convolutional Neural Networks \cite{luo2024moderntcn}, and Transformers \cite{wu2021autoformer,nie2022time,liu2023itransformer}, have shown promise in tackling these challenges. However, as sequence length increases, the number of model parameters grows rapidly, significantly heightening the risk of overfitting \cite{zeng2023transformers}.

\begin{figure}[t]
    \centering

    \noindent %
    \begin{minipage}[c]{0.28\textwidth}
        \begin{tabular}{rl}
            (a) & DLinear \cite{zeng2023transformers}\\
                & MSE=0.399\\
                & \#Params=\textbf{18K}
        \end{tabular}
        \label{fig:1a}
    \end{minipage}
    \hfill
    \begin{minipage}[c]{0.7\textwidth}
        \begin{minipage}[c]{0.05\textwidth}\centering\mbox{}\end{minipage}\hfill %
        \begin{minipage}[c]{0.2\textwidth}\centering\includegraphics[width=\textwidth]{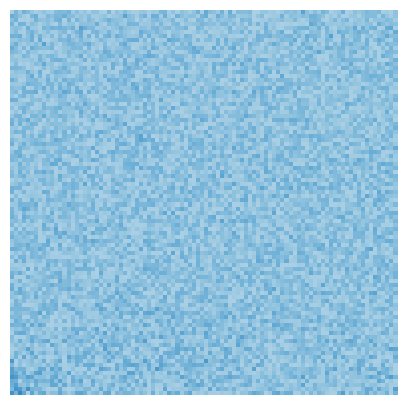}\end{minipage}\hfill
        \begin{minipage}[c]{0.05\textwidth}\centering\mbox{}\end{minipage}\hfill %
        \begin{minipage}[c]{0.2\textwidth}\centering\includegraphics[width=\textwidth]{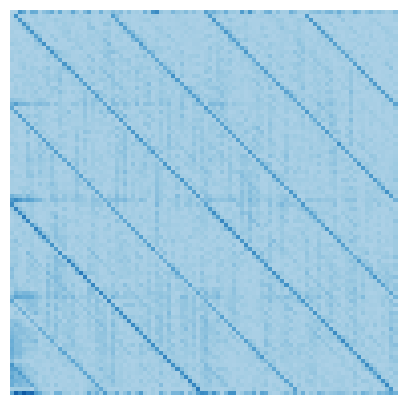}\end{minipage}\hfill
        \begin{minipage}[c]{0.05\textwidth}\centering\mbox{}\end{minipage}\hfill %
        \begin{minipage}[c]{0.1\textwidth}\centering$\longrightarrow$\end{minipage}\hfill
        \begin{minipage}[c]{0.2\textwidth}\centering\includegraphics[width=\textwidth]{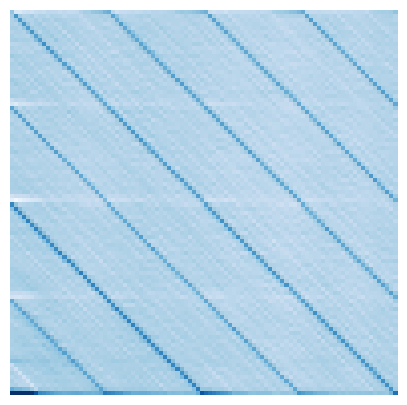}\end{minipage}
        \\
        \begin{minipage}[c]{0.05\textwidth}\centering\mbox{}\end{minipage}\hfill %
        \begin{minipage}[t]{0.2\textwidth}\centering{\scriptsize Trend}\end{minipage}\hfill
        \begin{minipage}[c]{0.02\textwidth}\centering\mbox{}\end{minipage}\hfill %
        \begin{minipage}[t]{0.2\textwidth}\centering{\scriptsize Seasonal}\end{minipage}\hfill
        \begin{minipage}[c]{0.08\textwidth}\centering\mbox{}\end{minipage}\hfill %
        \begin{minipage}[t]{0.2\textwidth}\centering\null\end{minipage}\hfill %
        \begin{minipage}[t]{0.1\textwidth}\centering\null\end{minipage}\hfill
        \begin{minipage}[t]{0.2\textwidth}\centering{\scriptsize IR}\end{minipage}
    \end{minipage}%

    \noindent
    
    \begin{minipage}[c]{0.3\textwidth}
        \begin{tabular}{rl}
            (b) & FITS\cite{xu2023fits}\\
                & MSE=0.385\\
                & \#Params=\textbf{6.4K}
        \end{tabular}
        \label{fig:1b}
    \end{minipage}
    \hfill
    \begin{minipage}[c]{0.69\textwidth}
        \begin{minipage}[c]{0.05\textwidth}\centering\mbox{}\end{minipage}\hfill %
        \begin{minipage}[c]{0.2\textwidth}\centering\includegraphics[width=\textwidth]{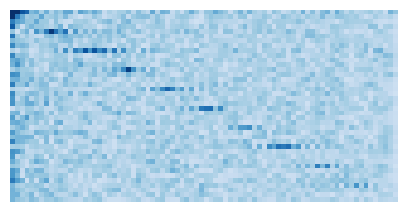}\end{minipage}\hfill
        \begin{minipage}[c]{0.05\textwidth}\centering\mbox{}\end{minipage}\hfill %
        \begin{minipage}[c]{0.2\textwidth}\centering\includegraphics[width=\textwidth]{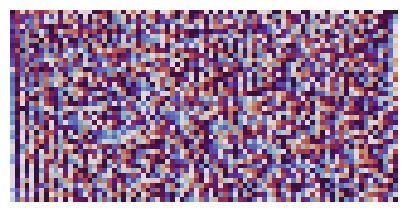}\end{minipage}\hfill
        \begin{minipage}[c]{0.05\textwidth}\centering\mbox{}\end{minipage}\hfill %
        \begin{minipage}[c]{0.1\textwidth}\centering$\longrightarrow$\end{minipage}\hfill
        \begin{minipage}[c]{0.2\textwidth}\centering\includegraphics[width=\textwidth]{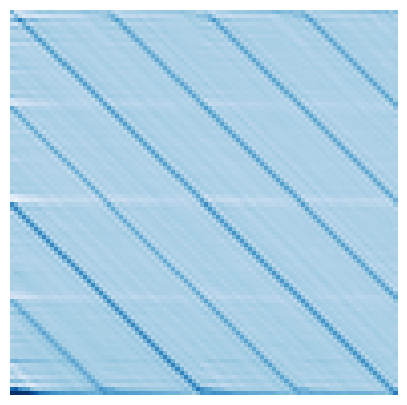}\end{minipage}
        \\
        \begin{minipage}[c]{0.05\textwidth}\centering\mbox{}\end{minipage}\hfill %
        \begin{minipage}[c]{0.2\textwidth}\centering{\scriptsize Amplitude}\end{minipage}\hfill
        \begin{minipage}[c]{0.05\textwidth}\centering\mbox{}\end{minipage}\hfill %
        \begin{minipage}[c]{0.2\textwidth}\centering{\scriptsize Phase}\end{minipage}\hfill
        \begin{minipage}[c]{0.05\textwidth}\centering\mbox{}\end{minipage}\hfill %
        \begin{minipage}[c]{0.1\textwidth}\centering\mbox{}\end{minipage}\hfill
        \begin{minipage}[c]{0.2\textwidth}\centering{\scriptsize IR}\end{minipage}
    \end{minipage}%

    \noindent
    \begin{minipage}[c]{0.3\textwidth}
        \begin{tabular}{rl}
            (c) & \ours\ (Ours)\\
                & MSE=0.384\\
                & \#Params=\textbf{0.7K}
        \end{tabular}
        \label{fig:1c}
    \end{minipage}
    \hfill
    \begin{minipage}[c]{0.69\textwidth}
        \begin{minipage}[c]{0.2\textwidth}\centering\includegraphics[width=\textwidth]{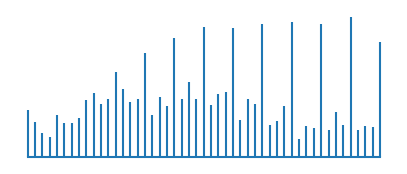}\end{minipage}\hfill
        \begin{minipage}[c]{0.2\textwidth}\centering\includegraphics[width=\textwidth]{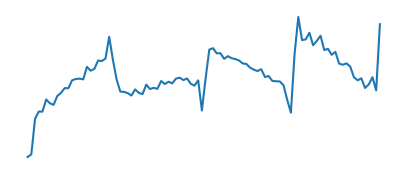}\end{minipage}\hfill
        \begin{minipage}[c]{0.2\textwidth}\centering\includegraphics[width=\textwidth]{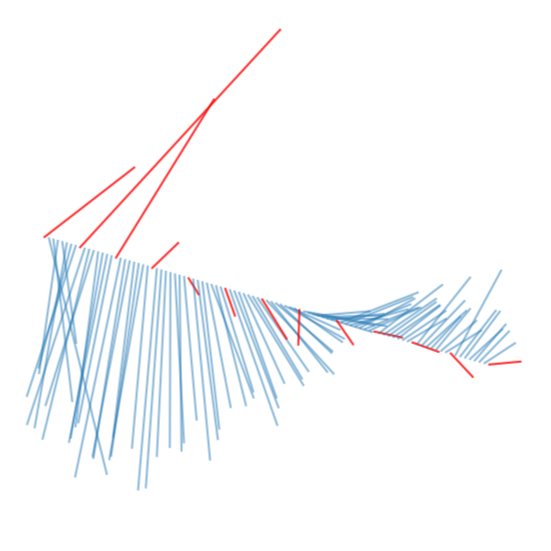}\end{minipage}\hfill
        \begin{minipage}[c]{0.1\textwidth}\centering$\longrightarrow$\end{minipage}\hfill
        \begin{minipage}[c]{0.2\textwidth}\centering\includegraphics[width=\textwidth]{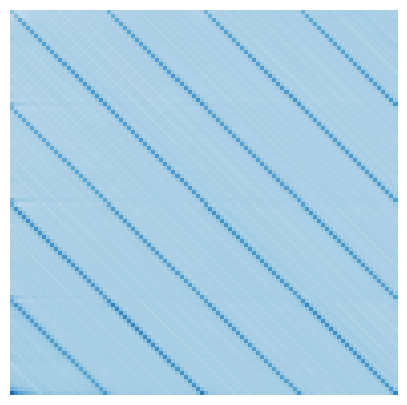}\end{minipage}
        \\
        \begin{minipage}[t]{0.2\textwidth}\centering{\scriptsize SFA}\end{minipage}\hfill
        \begin{minipage}[t]{0.2\textwidth}\centering{\scriptsize STA}\end{minipage}\hfill
        \begin{minipage}[t]{0.2\textwidth}\centering{\scriptsize IFM}\end{minipage}\hfill
        \begin{minipage}[t]{0.1\textwidth}\centering\null\end{minipage}\hfill
        \begin{minipage}[t]{0.2\textwidth}\centering{\scriptsize IR}\end{minipage}
    \end{minipage}%

    \caption{Weight visualization (left) and Impulse Response (IR) of equivalent Linear Time-Invariant system (right) of models trained on the ETTh1 dataset with look-back window $L=96$ and forecasting horizon $L^\prime=96$.}
    \label{fig:1}
\end{figure}

In response to this overfitting challenge, Linear Fully Connected (FC) models have gained renewed attention. Recent research has shown that such FC models can match or even outperform complex nonlinear models, while requiring substantially fewer parameters \cite{zeng2023transformers,li2023revisiting,xu2023fits,pmlr-v235-lin24n}. 
However, despite addressing the parameter excess of nonlinear models, these FC models are often {monolithic}. They attempt to learn complex temporal dynamics using a single, dense weight matrix $W \in \mathbb{R}^{L' \times L}$. This all-in-one design, which simultaneously captures frequential properties, temporal dependencies, and the input-output mapping, introduces its own critical issues: it not only leads to significant parameter redundancy, evidenced by its quadratic complexity, but also obscures the model's inner working mechanisms.

The inefficiency of this monolithic design can be observed directly. For instance, the weight matrix of a trained DLinear\cite{zeng2023transformers} model (Fig.~\ref{fig:1a}a) exhibits distinct horizontal and diagonal stripes. These stripes {represent the same pattern repeating at different periodic lags}, revealing a {massive parameter redundancy}—the model is inefficiently storing the same temporal dependency multiple times. Meanwhile, the FITS\cite{xu2023fits} off-diagonal amplitude and phase visualization (Fig.~\ref{fig:1b}b) appears noisy. These visualizations suggest that while FC models {attempt} to learn structural properties, they do so {implicitly} and {inefficiently}. We identify this core problem as {entanglement}: the properties are wastefully embedded within a single, large, dense matrix.

This motivates our core idea: to replace the monolithic $W$ matrix with an {explicit factorization}, or {disentanglement}, into a sequence of specialized, interpretable, and parameter-efficient linear modules. We disentangle the core temporal dynamics into three key priors: (1) \textbf{Frequential Filtering}: In recognition of the fact that time series data are typically decomposed into trend and seasonal components \cite{wu2021autoformer}, we incorporate the \textbf{Static Frequential Attention (SFA)} module to prioritize predictable frequencies. (2) \textbf{Temporal Importance}: Not all historical time steps are equally important. This motivates the \textbf{Static Time Attention (STA)} module to focus on pivotal time steps. (3) \textbf{Independent Mapping}: To model these periodicities holistically and efficiently through a time-invariant mapping (thus avoiding the explicit redundancy seen in Fig.~\ref{fig:1a}), we propose the \textbf{Independent Frequential Mapping (IFM)} approach. {Furthermore}, to efficiently capture inter-variable dependencies \cite{peiwen2023channelindependentstrategyoptimal} in the multivariate setting, we introduce a (4) \textbf{Low-rank Weight Sharing} policy. Our explicit factorization (Fig.~\ref{fig:1c}c) achieves a cleaner, more efficient representation, directly addressing the redundancy observed in Fig.~\ref{fig:1a}a and~\ref{fig:1a}b.

\begin{wrapfigure}{r}{0.4\textwidth}
    \centering
    \includegraphics[width=\linewidth]{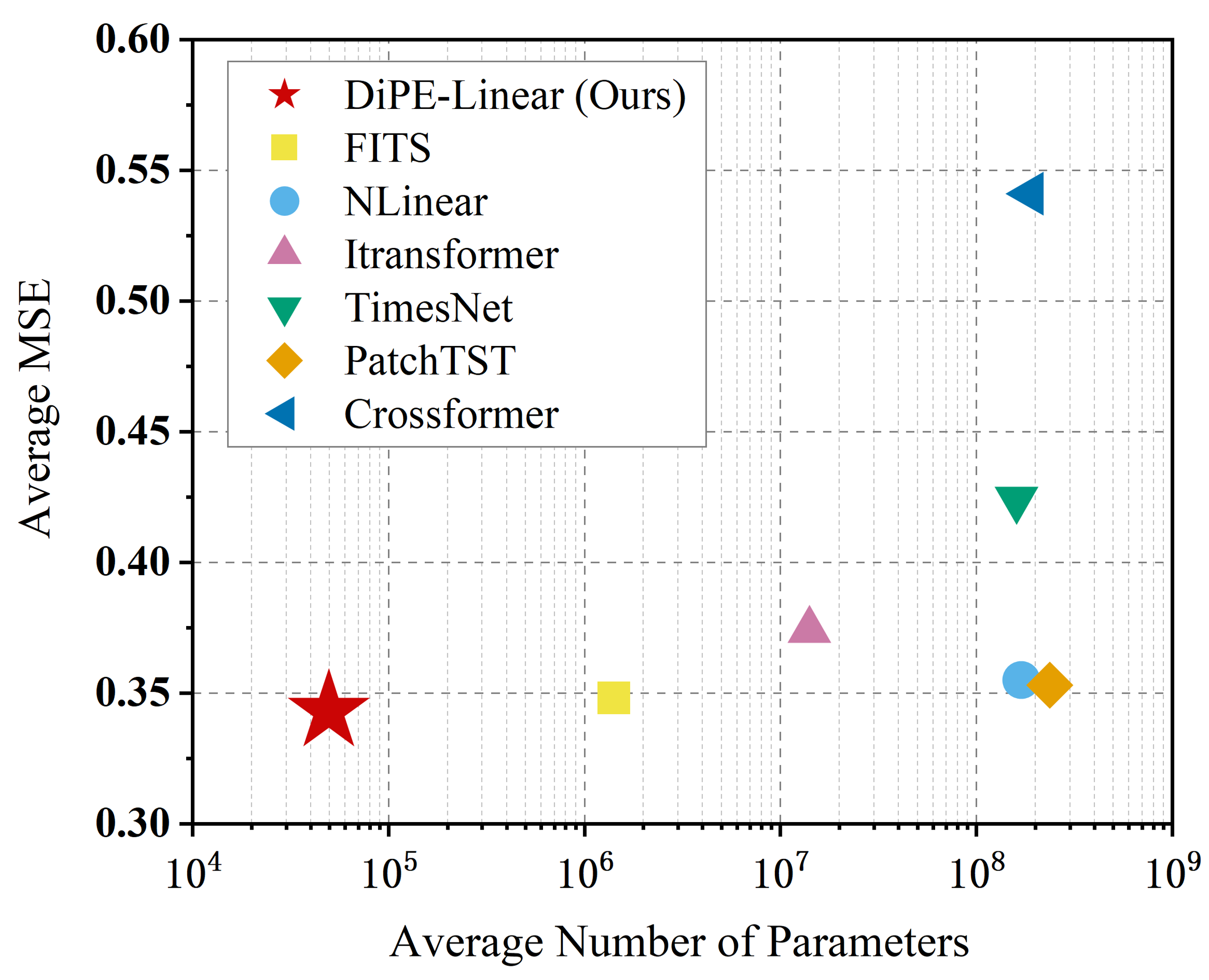}
    \caption{Mean Squared Error (MSE) and number of parameters compared with other models. 
    }
    \label{fig:mse-param}
\end{wrapfigure}

Building upon the above-mentioned motivation, we propose a novel \textbf{Di}sentangled \textbf{P}arameter-\textbf{E}fficient \textbf{Linear} model (\textbf{\ours}) for LTSF. In this work, "disentangled" refers to the model's architecture which factorizes the complex, monolithic FC weight matrix into a sequence of specialized, interpretable, and parameter-efficient linear modules (SFA, STA, and IFM). Each module addresses a distinct aspect of temporal modeling (frequential filtering, temporal importance weighting, and frequential mapping, respectively). These linear sub-modules collectively reduce the FC’s quadratic parameter complexity to linear complexity and quadratic computational complexity to log-linear. This lightweight, fully linear design achieves state-of-the-art performance with minimal parameters, as shown in Fig.~\ref{fig:1a} and \ref{fig:mse-param}, establishing an efficient and reliable pipeline for time series forecasting. 

In summary, our contributions are multifold:

\begin{itemize}
    \item We propose \ours, a novel disentangled parameter-efficient linear model for LTSF. Its innovative design, {motivated by an analysis of entanglement in existing FCs}, features specialized sub-modules that improve performance and {offer clear interpretability}.
    \item The proposed \ours~significantly reduces the parameter complexity (from quadratic to linear) as well as computational complexity (to log-linear) {without sacrificing accuracy}.
    \item Extensive experiments demonstrate that \ours~achieves performance comparable to or surpassing other state-of-the-art models, validating its effectiveness and efficiency.
\end{itemize}

\section{Related Work}

\subsubsection{Efficient Long-Term Forecasting.}
LTSF demands models that handle long sequences efficiently. Early Transformer-based models \cite{NEURIPS2019_6775a063,Kitaev2020Reformer,zhou2021informer,wu2021autoformer,pmlr-v162-zhou22g} reduced computational complexity from $\mathcal{O}(L^2)$ to $\mathcal{O}(L)$, but their complex structure leads to a large number of parameters, causing overfitting on long inputs \cite{zeng2023transformers}. DLinear \cite{zeng2023transformers} proposed a simpler, single-layer fully connected network with $\mathcal{O}(L^2)$ complexity, offering lower overall complexity than Transformer-based models, though it still suffers from parameter redundancy. FITS \cite{xu2023fits} reduced parameter count by applying low-pass filters in the frequency domain, while SparseTSF \cite{pmlr-v235-lin24n} introduced Cross-Period Sparse Forecasting to achieve further parameter reduction. However, both methods trade off accuracy to reduce parameters \cite{toner2024analysis,pmlr-v235-lin24n}, and addressing parameter redundancy in FC models remains an open challenge.

\subsubsection{Exploiting Frequency-Domain Characteristics.}
Recent advances in time series forecasting have leveraged frequency-domain techniques to enhance performance. FEDformer \cite{pmlr-v162-zhou22g} introduced a DFT-based frequency-enhanced attention mechanism, while FiLM \cite{zhou2022film} used Fourier analysis to retain historical information and reduce noise. Other methods, such as FreTS \cite{yi2023frequencydomain} and FITS \cite{xu2023fits}, work entirely in the frequency domain, with FreTS using a frequency-domain MLP for both temporal and channel-wise dependencies. Additionally, frequency-domain loss functions have been integrated into time-domain objectives. For example, FTMixer \cite{li2024ftmixerfrequencytimedomain} and FreDF \cite{wang2024fredflearningforecastfrequency} use frequency-domain loss components to improve alignment with the ground truth.

\subsubsection{Channel Dependence for Multivariate Forecasting.}
DLinear \cite{zeng2023transformers} introduced the Channel-Independent strategy for multivariate time series forecasting, which, despite ignoring causal relationships, led to notable performance gains. PatchTST \cite{nie2022time} further validated this approach in Transformer-based models, showing that independently processing variables while sharing model weights across them can be effective. RLinear \cite{li2023revisiting} demonstrated that using independent weights for variables with distinct periodicities can further improve performance. Subsequent research has sought an optimal balance between weight sharing and independence. MLinear \cite{li2023mlinearrethinklinearmodel} dynamically adjusts strategies based on temporal semantics, while CSC \cite{peiwen2023channelindependentstrategyoptimal} reassigns weight-variable mappings during validation to cluster variables. However, MLinear introduces extra computational overhead, and CSC relies on assumptions about weight sharing that may not always hold.

\section{Method}

\begin{figure*}[thb]
    \centering
    \includegraphics[width=\textwidth]{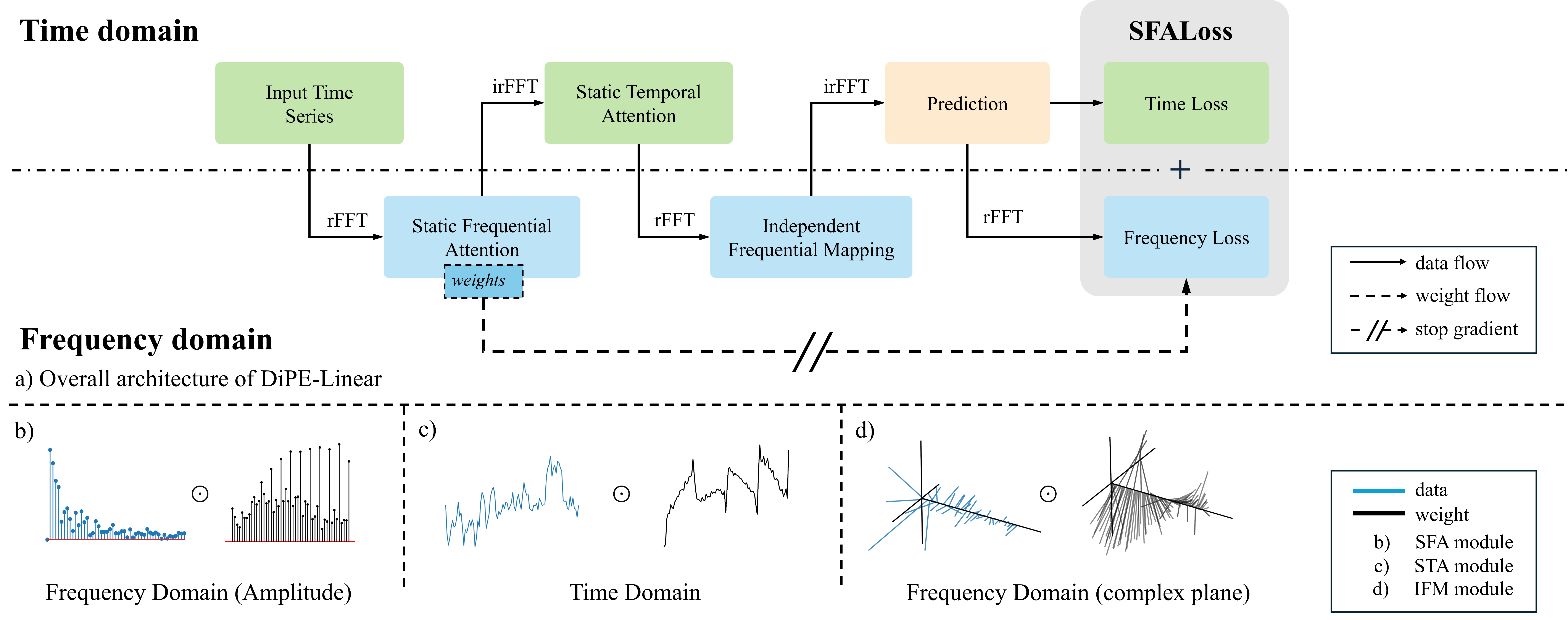}
    \caption{Overall architecture of \ours. }
    \label{fig:3}
\end{figure*}

We first define the multivariate LTSF problem, then describe the \ours\ architecture, focusing on its three core modules: Static Frequential Attention (SFA), Static Temporal Attention (STA), and Independent Frequential Mapping (IFM), assuming univariate time series. We then explain how the Low-rank Weight Sharing method extends these modules to multivariate time series. Finally, we introduce SFALoss, a loss function that combines weighted losses in both frequency and time domains to prioritize predictable frequencies and improve model performance.

\subsection{Problem Definition}
The multivariate LTSF problem is formally defined as follows: Given a historical time series $\mathbf{x} \in \mathbb{R}^{C \times L}$, where $C$ denotes the number of variates and $L$ represents the look-back length, the goal is to predict the future time series $\mathbf{y} \in \mathbb{R}^{C \times L^\prime}$, where $L^\prime$ is the forecasting horizon.

\subsection{Static Frequential Attention}

We introduce the SFA module as a frequency-domain filter to enable the model to selectively enhance and amplify relevant frequencies. Additionally, to preserve the temporal structure of the series and avoid interference with subsequent temporal feature extraction, we constrained this filter to be a zero-phase filter. The SFA operates by first applying real Fast
Fourier Transform (rFFT) to transform the time-domain data into the frequency domain. In this domain, a learned filter is applied via element-wise multiplication, selectively amplifying or suppressing specific frequency components. As a zero-phase filter, it exclusively modifies the amplitude of the signal while preserving the original phase information. Finally, inverse
rFFT (irFFT) is used to reverse the filtered data back to the time domain. Specifically, the enhanced signal $\mathbf{z}_{\text{SFA}}$ is given by

\begin{equation}
\label{eq:SFA}
    \mathbf{z}_{\text{SFA}} = \invfft (\theta_{\text{SFA}} \odot \fft(\mathbf{x})),
\end{equation}
where $\mathbf{x} \in \mathbb{R}^{L}$ represents the input univariate series
, $\fft$ denotes the rFFT, $\invfft$ is the irFFT, $\theta_{\text{SFA}} \in \mathbb{R}^{\lfloor {L}/{2} \rfloor+1}$ is the learnable static frequential attention map,  and $\odot$ denotes element-wise multiplication.

\subsection{Static Temporal Attention}

We apply SFA {before} STA based on the hypothesis that the model should first isolate the most predictable, core frequencies of the signal (SFA), and {then} identify the critical time steps within that filtered signal (STA). Reversing the order might cause the STA module to focus on temporally salient but high-frequency noise.

Following this, to enable the model to have the capability of capturing important input time points in the temporal domain, we introduce the STA as a second component of our model. This module enables the input time series to undergo element-wise multiplication with a learned temporal attention map, effectively assigning appropriate importance to relevant historical time points and thereby enhancing the model's ability to capture temporal dependencies. The processed signal $\mathbf{z}_{\text{STA}}$ is thus given by

\begin{equation}
\label{eq:STA}
    \mathbf{z}_{\text{STA}} = \theta_{\text{STA}}\odot \mathbf{z}_{\text{SFA}},
\end{equation}
where $\theta_{\text{STA}} \in \mathbb{R}^{L}$ is the learnable time-domain attention map.

\subsection{Independent Frequential Mapping}

While the SFA and STA modules are part of the input series pre-processing, the IFM module directly maps the input historical series to the forecasted output under the assumption of independence between different frequencies.

We acknowledge that this assumption is a theoretical simplification, as real-world signals often exhibit complex inter-frequency dependencies (e.g., harmonics \cite{wang2024fredflearningforecastfrequency}). However, we adopt this assumption as a deliberate design choice. Explicitly modeling these intricate dependencies within a linear framework could increase the risk of multicollinearity and overfitting.

Therefore, our approach acts as an effective {inductive bias}. By enforcing frequency-wise independence, the IFM module learns a 1D convolution (Eq. \ref{eq:6}) that is inherently {time-invariant} and {phase-preserving} (relative to each frequency component), which are desirable properties for modeling stationary seasonal patterns. This compels the model to focus on the primary predictive information within each frequency component rather than capturing potentially spurious cross-frequency correlations. As our extensive experiments on diverse datasets (see Section \ref{sec:experiments}) demonstrate, this design choice consistently leads to state-of-the-art performance, suggesting it enhances model generalization and robustness rather than limiting its accuracy.

Following this assumption, for each frequency component, we apply a complex-valued multiply-accumulate operation independently. To transform the input space $\mathbb{R}^{L}$ to the output space $\mathbb{R}^{L^\prime}$, we apply zero-padding to the input series, extending its length to $L + L^\prime$ before performing the rFFT. At the end, the last $L^\prime$ values are extracted and used as the predicted output. The forecasted output $\hat{\mathbf{y}}$ is thus defined as:

\begin{equation}
\label{eq:IFM1}
    \mathbf{Z}_{\text{STA\_pad}} = \fft (\texttt{Zero\_Padding}(\mathbf{z}_{\text{STA}}))
\end{equation}

\begin{equation}
\label{eq:IFM2}
    \hat{\mathbf{Y}}_\text{pad} =\theta_{\text{IFM}}\odot \mathbf{Z}_{\text{STA\_pad}}  + \beta_{\text{IFM}}
\end{equation}

\begin{equation}
\label{eq:IFM3}
    \hat{\mathbf{y}} = \invfft (\hat{\mathbf{Y}}_\text{pad})_{\left[-L^\prime:\right]},
\end{equation}
where $\theta_{\text{IFM}} \in \mathbb{C}^{\lfloor(L + L^\prime - 1)/2\rfloor+1}$ is the complex-valued weight of the IFM module, and $\beta_{\text{IFM}} \in \mathbb{C}^{\lfloor(L + L^\prime - 1)/2\rfloor+1}$ is the complex-valued bias.

According to the Convolution Theorem, this process can be equivalently represented as a 1D convolution in the time domain:

\begin{equation}
    \hat{\mathbf{y}} = \invfft (\theta_{\text{IFM}})  \ast \mathbf{z}_{\text{STA}} + \invfft (\beta_{\text{IFM}}),
    \label{eq:6}
\end{equation}
where $\invfft (\theta_{\text{IFM}}) \in \mathbb{R}^{L + L^\prime - 1}$ is the convolution kernel in time domain. The extremely large convolution kernel distinguishes it from traditional CNNs that use much smaller kernels. This larger kernel allows each output to capture a global receptive field, enabling the model to incorporate information from the entire input sequence.

\subsection{Low-rank Weight Sharing}

Weight sharing is commonly used in time series forecasting models \cite{zeng2023transformers,xu2023fits,nie2022time} under the assumption that different channels exhibit similar patterns. However, this assumption does not hold in datasets like Weather, where variables such as temperature, pressure, and rainfall follow distinct patterns. While training weights independently for each channel can improve performance \cite{li2023revisiting}, it significantly increases model complexity. To address this, we propose selecting a lower rank based on pattern similarities, reducing the number of weight sets and balancing efficiency with performance. Inspired by mixture-of-experts \cite{pmlr-v238-ni24a} and dynamic convolution \cite{Chen_2020_CVPR}, we introduce a novel Low-rank Weight Sharing architecture, optimizing parameter efficiency and predictive performance.

Specifically, for a multivariate time series with $C$ variables, we introduce a hyperparameter $M$, where $M$ denotes the number of independent weight sets to be learned, with $M \ll C$. For each component, $M$ distinct sets of weights are learned, alongside a learnable routing matrix. The low-rank routing matrix $\mathbf{R} \in \mathbb{R}^{M\times C}$ is a model parameter designed to linearly combine the weights and assign each variable to an appropriate weight set. We regularize this routing matrix using the Softmax function, where $\tau$ is the Softmax temperature:

\begin{equation}
    \mathbf{R} ^\prime = \text{Softmax}(\frac{\mathbf{R}}{\tau}).
\end{equation}

We denote $\{\mathcal{G}_m\}_{m \in \llbracket 1, M \rrbracket}$ as the set of mappings representing any given module, corresponding to the $M$ independent weight sets learned. For variate $c \in \llbracket 1, C \rrbracket$, the weight $\mathcal G_c^\prime$ actually applied to channel $c$ is:

\begin{equation}
    \mathcal G_c^\prime = \sum_{m=1}^M \mathbf R ^\prime _c \cdot \mathcal G _m.
\end{equation}

\subsection{SFALoss}

Building on the insights from FreDF \cite{wang2024fredflearningforecastfrequency}, we introduce SFALoss (denoted as $\mathcal{L}$), which integrates the Weighted Mean Absolute Error (WMAE) in the frequency domain ($\mathcal{L}_F$) and Mean Squared Error (MSE) in the time domain ($\mathcal{L}_T$). We empirically found this combination to be most effective: WMAE for $\mathcal{L}_F$ provides a more robust optimization signal that is less sensitive to large magnitude errors in a few dominant frequencies, while the standard MSE for $\mathcal{L}_T$ ensures overall predictive accuracy in the time domain. The frequency-domain loss ($\mathcal{L}_F$) is modulated by element-wise multiplication with the SFA weighting factor, $\theta_{\text{SFA}}$, allowing prioritizing relevant frequencies:

\begin{equation}
    \mathcal{L}_F = \frac{1}{C} \sum_{c=1}^C \frac{\langle  \mathbf{R}^\prime_c \cdot \mathbf \theta_{\text{SFA}} , | \mathbf{Y}_{c} - \hat{\mathbf{Y}}_{c} |\rangle}{\lVert \mathbf{R}^\prime_c \cdot \theta_{\text{SFA}} \rVert_1} ,
\end{equation}
where $\mathbf{Y} = \fft(\mathbf{y})$ and $\hat{\mathbf{Y}} = \fft(\hat{\mathbf{y}})$ are respectively ground truth and predicted future time series in frequency-domain. While computing the loss, $\theta_{\text{SFA}}$ is detached from the computational graph, ensuring that no gradients are backpropagated through it. This crucial step prevents a trivial solution where the SFA module might learn to suppress (i.e., set weights to zero for) frequencies that are difficult to predict, rather than learning the frequencies most relevant to the primary time-domain forecasting task.

We also preserve the MSE loss in time domain $\mathcal{L}_T$:

\begin{equation}
    \mathcal{L}_T = \frac{1}{C} \frac{1}{L^\prime} \sum_{c=1}^C \sum_{i=1}^{L^\prime} \|\hat{y}_{c,i} - y_{c,i}\|_2^2.
\end{equation}

Finally, the overall SFAloss function is defined as follow:

\begin{equation}
    \mathcal{L} = \alpha \mathcal{L}_F + (1 - \alpha) \mathcal{L}_T,
\end{equation}
where $\alpha \in [0, 1]$ is hyperparameter that balances the contribution of the two loss terms, allowing us to achieve a trade-off between them.

\section{Experiment}\label{sec:experiments}

\subsection{Experimental Setup}

\subsubsection{Datasets.}
We evaluate our model on 10 real-world datasets, comprising 8 for Long-term (LTSF) and 2 for Short-term (STSF) forecasting. The LTSF datasets include \textbf{ETT} (ETTh1, ETTh2, ETTm1, ETTm2) \cite{zhou2021informer}, \textbf{Weather} \cite{zhou2021informer}, \textbf{Electricity}, and two private cloud service datasets (\textbf{IaaS}, \textbf{FaaS}). The STSF datasets are \textbf{M5} and \textbf{Illness} \cite{wu2021autoformer}. Further preprocessing details are available in the accompanying code.

\subsubsection{Implementation Details.}
Models are trained for 50 epochs with a $0.001$ learning rate. Batch sizes are set to 64 (LTSF) and 8 (STSF). Forecasting horizons $L^\prime$ are $\{96, 192, 360, 720\}$ for LTSF and $\{24, 36, 48, 60\}$ for STSF. Input sequence lengths $L$ are 720 (public LTSF), 2880 (private LTSF), and 60 (STSF). We set $M=4$ for Electricity and Weather, $M=1$ otherwise, and apply linear annealing for the softmax temperature \cite{Chen_2020_CVPR}.

\subsubsection{Baselines and Metrics.}
We benchmark our approach against leading linear models, including FITS \cite{xu2023fits}, RLinear \cite{li2023revisiting}, DLinear, and NLinear \cite{zeng2023transformers}. We also compare with established non-linear models (e.g., iTransformer \cite{liu2023itransformer}, PatchTST \cite{dosovitskiy2021an}, TimesNet \cite{wu2023timesnet}) and lightweight baselines (SparseTSF \cite{pmlr-v235-lin24n}, FITS \cite{xu2023fits}). Performance is evaluated using MSE and MAE.

\subsection{Main Results}

\subsubsection{Versus FC Models.}

\begin{table}[htb!]

\caption{Results compared with LFCs. Average and standard deviation($\pm$) are calculated over five runs. The best and second results are highlighted in \best{bold} and \second{underline}.}
\label{tab:2}
\centering
\resizebox{\textwidth}{!}{
\begin{tabular}{cccccccccccc}
\toprule
 &
   &
  \multicolumn{2}{c}{\textbf{\ours~(Ours)}} &
  \multicolumn{2}{c}{FITS (2024)} &
  \multicolumn{2}{c}{RLinear (2023)} &
  \multicolumn{2}{c}{DLinear (2023)} &
  \multicolumn{2}{c}{Nlinear (2023)} \\
\midrule
 \multicolumn{2}{c}{Metrics $\downarrow$} &
  MSE &
  MAE &
  MSE &
  MAE &
  MSE &
  MAE &
  MSE &
  MAE &
  MSE &
  MAE \\
\midrule
\multirow{4}{*}{ETTh1} &
  96 &
  \best{0.369}\tiny$\pm$0.000 &
  \best{0.393}\tiny$\pm$0.000 &
  0.380\tiny$\pm$0.000 &
  0.403\tiny$\pm$0.000 &
  \second{0.376}\tiny$\pm$0.004 &
  \second{0.401}\tiny$\pm$0.004 &
  0.388\tiny$\pm$0.010 &
  0.414\tiny$\pm$0.012 &
  0.384\tiny$\pm$0.004 &
  0.405\tiny$\pm$0.003 \\
 &
  192 &
  \best{0.407}\tiny$\pm$0.000 &
  \best{0.415}\tiny$\pm$0.000 &
  0.417\tiny$\pm$0.001 &
  0.425\tiny$\pm$0.001 &
  0.416\tiny$\pm$0.002 &
  0.426\tiny$\pm$0.002 &
  0.425\tiny$\pm$0.005 &
  0.436\tiny$\pm$0.005 &
  \second{0.415}\tiny$\pm$0.003 &
  \second{0.424}\tiny$\pm$0.002 \\
 &
  336 &
  \best{0.424}\tiny$\pm$0.000 &
  \best{0.427}\tiny$\pm$0.000 &
  \second{0.436}\tiny$\pm$0.000 &
  \second{0.440}\tiny$\pm$0.000 &
  0.444\tiny$\pm$0.002 &
  0.444\tiny$\pm$0.002 &
  0.469\tiny$\pm$0.010 &
  0.469\tiny$\pm$0.009 &
  0.445\tiny$\pm$0.002 &
  0.443\tiny$\pm$0.002 \\
 &
  720 &
  \best{0.409}\tiny$\pm$0.000 &
  \best{0.439}\tiny$\pm$0.000&
  \second{0.432}\tiny$\pm$0.000 &
  \second{0.456}\tiny$\pm$0.000 &
  0.474\tiny$\pm$0.002 &
  0.481\tiny$\pm$0.002 &
  0.530\tiny$\pm$0.021 &
  0.533\tiny$\pm$0.015 &
  0.439\tiny$\pm$0.002 &
  0.470\tiny$\pm$0.000 \\
\midrule
\multirow{4}{*}{ETTh2} &
  96 &
  0.275\tiny$\pm$0.001 &
  \second{0.336}\tiny$\pm$0.001 &
  \second{0.272}\tiny$\pm$0.000 &
  \second{0.336}\tiny$\pm$0.000 &
  \best{0.270}\tiny$\pm$0.001 &
  \best{0.335}\tiny$\pm$0.001 &
  0.280\tiny$\pm$0.005 &
  0.345\tiny$\pm$0.004 &
  0.276\tiny$\pm$0.002 &
  0.338\tiny$\pm$0.001 \\
 &
  192 &
  \best{0.325}\tiny$\pm$0.001 &
  \best{0.372}\tiny$\pm$0.000 &
  \second{0.331}\tiny$\pm$0.000 &
  \second{0.374}\tiny$\pm$0.000 &
  0.335\tiny$\pm$0.004 &
  0.380\tiny$\pm$0.002 &
  0.358\tiny$\pm$0.014 &
  0.399\tiny$\pm$0.011 &
  0.345\tiny$\pm$0.004 &
  0.382\tiny$\pm$0.002 \\
 &
  336 &
  \best{0.350}\tiny$\pm$0.002 &
  \best{0.393}\tiny$\pm$0.001 &
  \second{0.354}\tiny$\pm$0.000 &
  \second{0.395}\tiny$\pm$0.000 &
  0.366\tiny$\pm$0.005 &
  0.409\tiny$\pm$0.003 &
  0.439\tiny$\pm$0.017 &
  0.457\tiny$\pm$0.009 &
  0.375\tiny$\pm$0.009 &
  0.411\tiny$\pm$0.004 \\
 &
  720 &
  \best{0.375}\tiny$\pm$0.002 &
  \best{0.415}\tiny$\pm$0.001 &
  \second{0.378}\tiny$\pm$0.000 &
  \second{0.423}\tiny$\pm$0.000 &
  0.414\tiny$\pm$0.003 &
  0.447\tiny$\pm$0.001 &
  0.657\tiny$\pm$0.062 &
  0.573\tiny$\pm$0.026 &
  0.408\tiny$\pm$0.008 &
  0.446\tiny$\pm$0.003 \\
\midrule
\multirow{4}{*}{ETTm1} &
  96 &
  \best{0.309}\tiny$\pm$0.000 &
  \best{0.350}\tiny$\pm$0.000 &
  \best{0.309}\tiny$\pm$0.000 &
  \second{0.352}\tiny$\pm$0.001 &
  \best{0.309}\tiny$\pm$0.001 &
  \second{0.352}\tiny$\pm$0.001 &
  \second{0.312}\tiny$\pm$0.005 &
  0.358\tiny$\pm$0.006 &
  0.318\tiny$\pm$0.007 &
  0.357\tiny$\pm$0.005 \\
 &
  192 &
  \best{0.339}\tiny$\pm$0.000 &
  \best{0.369}\tiny$\pm$0.000 &
  \best{0.339}\tiny$\pm$0.001 &
  \best{0.369}\tiny$\pm$0.001 &
  \second{0.341}\tiny$\pm$0.004 &
  \second{0.370}\tiny$\pm$0.003 &
  0.350\tiny$\pm$0.006 &
  0.386\tiny$\pm$0.008 &
  0.350\tiny$\pm$0.007 &
  0.377\tiny$\pm$0.005 \\
 &
  336 &
  \best{0.367}\tiny$\pm$0.000 &
  \best{0.386}\tiny$\pm$0.000 &
  \second{0.368}\tiny$\pm$0.001 &
  \best{0.386}\tiny$\pm$0.001 &
  0.369\tiny$\pm$0.003 &
  \second{0.387}\tiny$\pm$0.003 &
  0.379\tiny$\pm$0.005 &
  0.403\tiny$\pm$0.006 &
  0.378\tiny$\pm$0.005 &
  0.393\tiny$\pm$0.004 \\
 &
  720 &
  \second{0.416}\tiny$\pm$0.000 &
  \second{0.413}\tiny$\pm$0.000 &
  \second{0.416}\tiny$\pm$0.000 &
  \second{0.413}\tiny$\pm$0.000 &
  \best{0.415}\tiny$\pm$0.001 &
  \best{0.411}\tiny$\pm$0.001 &
  0.439\tiny$\pm$0.011 &
  0.443\tiny$\pm$0.012 &
  0.420\tiny$\pm$0.003 &
  0.415\tiny$\pm$0.002 \\
\midrule
\multirow{4}{*}{ETTm2} &
  96 &
  \best{0.162}\tiny$\pm$0.000 &
  \second{0.252}\tiny$\pm$0.000 &
  \best{0.162}\tiny$\pm$0.000 &
  0.253\tiny$\pm$0.000 &
  \best{0.162}\tiny$\pm$0.000 &
  \best{0.251}\tiny$\pm$0.000 &
  \second{0.163}\tiny$\pm$0.001 &
  0.255\tiny$\pm$0.001 &
  \best{0.162}\tiny$\pm$0.000 &
  \second{0.252}\tiny$\pm$0.001 \\
 &
  192 &
  \best{0.216}\tiny$\pm$0.000 &
  \best{0.289}\tiny$\pm$0.000 &
  \best{0.216}\tiny$\pm$0.000 &
  0.291\tiny$\pm$0.000 &
  \best{0.216}\tiny$\pm$0.000 &
  \second{0.290}\tiny$\pm$0.000 &
  \second{0.219}\tiny$\pm$0.002 &
  0.297\tiny$\pm$0.001 &
  \best{0.216}\tiny$\pm$0.000 &
  0.291\tiny$\pm$0.000 \\
 &
  336 &
  \best{0.268}\tiny$\pm$0.000 &
  \best{0.324}\tiny$\pm$0.000 &
  \best{0.268}\tiny$\pm$0.000 &
  \second{0.326}\tiny$\pm$0.000 &
  \best{0.268}\tiny$\pm$0.001 &
  \second{0.326}\tiny$\pm$0.000 &
  \second{0.272}\tiny$\pm$0.003 &
  0.334\tiny$\pm$0.003 &
  \best{0.268}\tiny$\pm$0.000 &
  \second{0.326}\tiny$\pm$0.000 \\
 &
  720 &
  \second{0.353}\tiny$\pm$0.000 &
  \second{0.379}\tiny$\pm$0.000 &
  \best{0.349}\tiny$\pm$0.000 &
  \best{0.378}\tiny$\pm$0.000 &
  0.354\tiny$\pm$0.001 &
  0.384\tiny$\pm$0.000 &
  0.367\tiny$\pm$0.004 &
  0.398\tiny$\pm$0.003 &
  \best{0.349}\tiny$\pm$0.000 &
  \second{0.379}\tiny$\pm$0.000 \\
\midrule
\multirow{4}{*}{Electricity} &
  96 &
  \best{0.132}\tiny$\pm$0.000 &
  \best{0.228}\tiny$\pm$0.001 &
  0.134\tiny$\pm$0.000 &
  0.231\tiny$\pm$0.000 &
  0.135\tiny$\pm$0.000 &
  0.232\tiny$\pm$0.000 &
  \second{0.133}\tiny$\pm$0.000 &
  \second{0.229}\tiny$\pm$0.000 &
  \second{0.133}\tiny$\pm$0.000 &
  \best{0.228}\tiny$\pm$0.000 \\
 &
  192 &
  \best{0.148}\tiny$\pm$0.000 &
  0.245\tiny$\pm$0.000 &
  \second{0.149}\tiny$\pm$0.000 &
  0.244\tiny$\pm$0.000 &
  0.150\tiny$\pm$0.000 &
  0.246\tiny$\pm$0.000 &
  \best{0.148}\tiny$\pm$0.000 &
  \second{0.243}\tiny$\pm$0.000 &
  \best{0.148}\tiny$\pm$0.000 &
  \best{0.242}\tiny$\pm$0.000 \\
 &
  336 &
  \best{0.162}\tiny$\pm$0.000 &
  0.261\tiny$\pm$0.000 &
  0.165\tiny$\pm$0.000 &
  \second{0.260}\tiny$\pm$0.000 &
  0.166\tiny$\pm$0.000 &
  0.262\tiny$\pm$0.000 &
  \best{0.162}\tiny$\pm$0.000 &
  0.261\tiny$\pm$0.000 &
  \second{0.164}\tiny$\pm$0.000 &
  \best{0.258}\tiny$\pm$0.000 \\
 &
  720 &
  \second{0.198}\tiny$\pm$0.002 &
  0.296\tiny$\pm$0.001 &
  0.204\tiny$\pm$0.000 &
  \second{0.293}\tiny$\pm$0.000 &
  0.206\tiny$\pm$0.000 &
  0.294\tiny$\pm$0.000 &
  \best{0.196}\tiny$\pm$0.000 &
  \second{0.293}\tiny$\pm$0.001 &
  0.203\tiny$\pm$0.000 &
  \best{0.291}\tiny$\pm$0.000 \\
\midrule
\multirow{4}{*}{Weather} &
  96 &
  \best{0.142}\tiny$\pm$0.001 &
  0.201\tiny$\pm$0.001 &
  \best{0.142}\tiny$\pm$0.000 &
  \best{0.192}\tiny$\pm$0.000 &
  \second{0.143}\tiny$\pm$0.000 &
  \second{0.194}\tiny$\pm$0.000 &
  \best{0.142}\tiny$\pm$0.000 &
  0.202\tiny$\pm$0.000 &
  \best{0.142}\tiny$\pm$0.001 &
  \best{0.192}\tiny$\pm$0.001 \\
 &
  192 &
  0.187\tiny$\pm$0.003 &
  0.253\tiny$\pm$0.002 &
  0.185\tiny$\pm$0.000 &
  \best{0.234}\tiny$\pm$0.000 &
  0.186\tiny$\pm$0.000 &
  \second{0.235}\tiny$\pm$0.000 &
  \second{0.184}\tiny$\pm$0.001 &
  0.249\tiny$\pm$0.002 &
  \best{0.183}\tiny$\pm$0.000 &
  \best{0.234}\tiny$\pm$0.000 \\
 &
  336 &
  \best{0.234}\tiny$\pm$0.001 &
  0.293\tiny$\pm$0.002 &
  \second{0.235}\tiny$\pm$0.000 &
  \second{0.276}\tiny$\pm$0.000 &
  0.236\tiny$\pm$0.000 &
  \best{0.275}\tiny$\pm$0.000 &
  0.236\tiny$\pm$0.001 &
  0.293\tiny$\pm$0.001 &
  \best{0.234}\tiny$\pm$0.001 &
  \second{0.276}\tiny$\pm$0.000 \\
 &
  720 &
  \best{0.306}\tiny$\pm$0.004 &
  0.348\tiny$\pm$0.003 &
  \second{0.307}\tiny$\pm$0.000 &
  \second{0.328}\tiny$\pm$0.000 &
  0.308\tiny$\pm$0.000 &
  \best{0.327}\tiny$\pm$0.000 &
  \best{0.306}\tiny$\pm$0.002 &
  0.349\tiny$\pm$0.002 &
  0.308\tiny$\pm$0.000 &
  0.330\tiny$\pm$0.000 \\
\midrule
\multirow{4}{*}{FaaS} &
  96 &
  \best{0.280}\tiny$\pm$0.000 &
  \best{0.251}\tiny$\pm$0.000 &
  0.309\tiny$\pm$0.000 &
  0.266\tiny$\pm$0.000 &
  0.306\tiny$\pm$0.000 &
  \second{0.264}\tiny$\pm$0.000 &
  \second{0.303}\tiny$\pm$0.000 &
  \second{0.264}\tiny$\pm$0.000 &
  0.305\tiny$\pm$0.000 &
  0.266\tiny$\pm$0.000 \\
 &
  192 &
  \best{0.314}\tiny$\pm$0.000 &
  \best{0.279}\tiny$\pm$0.000 &
  0.342\tiny$\pm$0.000 &
  0.293\tiny$\pm$0.000 &
  0.344\tiny$\pm$0.000 &
  \second{0.292}\tiny$\pm$0.000 &
  \second{0.339}\tiny$\pm$0.000 &
  0.294\tiny$\pm$0.000 &
  0.340\tiny$\pm$0.000 &
  \second{0.292}\tiny$\pm$0.000 \\
 &
  336 &
  \best{0.351}\tiny$\pm$0.000 &
  \best{0.309}\tiny$\pm$0.000 &
  0.368\tiny$\pm$0.000 &
  0.316\tiny$\pm$0.000 &
  0.375\tiny$\pm$0.000 &
  0.319\tiny$\pm$0.000 &
  0.367\tiny$\pm$0.000 &
  0.323\tiny$\pm$0.000 &
  \second{0.364}\tiny$\pm$0.000 &
  \second{0.314}\tiny$\pm$0.000 \\
 &
  720 &
  \best{0.379}\tiny$\pm$0.000 &
  \best{0.332}\tiny$\pm$0.000 &
  0.415\tiny$\pm$0.000 &
  0.348\tiny$\pm$0.000 &
  0.436\tiny$\pm$0.000 &
  0.350\tiny$\pm$0.000 &
  0.417\tiny$\pm$0.000 &
  0.364\tiny$\pm$0.001 &
  \second{0.410}\tiny$\pm$0.000 &
  \second{0.344}\tiny$\pm$0.000 \\
\midrule
\multirow{4}{*}{IaaS} &
  96 &
  \best{0.789}\tiny$\pm$0.000 &
  \best{0.667}\tiny$\pm$0.000 &
  0.799\tiny$\pm$0.000 &
  0.702\tiny$\pm$0.000 &
  0.797\tiny$\pm$0.000 &
  0.688\tiny$\pm$0.000 &
  \second{0.795}\tiny$\pm$0.000 &
  0.699\tiny$\pm$0.000 &
  0.799\tiny$\pm$0.000 &
  \second{0.682}\tiny$\pm$0.000 \\
 &
  192 &
  \best{0.817}\tiny$\pm$0.000 &
  \best{0.698}\tiny$\pm$0.000 &
  0.839\tiny$\pm$0.000 &
  0.723\tiny$\pm$0.000 &
  \second{0.831}\tiny$\pm$0.000 &
  \second{0.700}\tiny$\pm$0.000 &
  0.823\tiny$\pm$0.000 &
  0.717\tiny$\pm$0.000 &
  0.820\tiny$\pm$0.000 &
  0.702\tiny$\pm$0.000 \\
 &
  336 &
  1.180\tiny$\pm$0.015 &
  0.839\tiny$\pm$0.010 &
  0.998\tiny$\pm$0.007 &
  0.855\tiny$\pm$0.005 &
  1.255\tiny$\pm$0.012 &
  0.822\tiny$\pm$0.007 &
  \best{0.842}\tiny$\pm$0.000 &
  \best{0.731}\tiny$\pm$0.000 &
  \second{0.898}\tiny$\pm$0.000 &
  \second{0.771}\tiny$\pm$0.000 \\
 &
  720 &
  \best{0.869}\tiny$\pm$0.000 &
  \best{0.736}\tiny$\pm$0.000 &
  0.899\tiny$\pm$0.000 &
  0.839\tiny$\pm$0.000 &
  0.938\tiny$\pm$0.000 &
  0.828\tiny$\pm$0.000 &
  \second{0.881}\tiny$\pm$0.000 &
  \second{0.741}\tiny$\pm$0.000 &
  0.924\tiny$\pm$0.000 &
  0.785\tiny$\pm$0.000 \\
\midrule
\multirow{4}{*}{Illness} &
  24 &
  \second{2.260}\tiny$\pm$0.000 &
  \second{0.951}\tiny$\pm$0.000 &
  2.289\tiny$\pm$0.000 &
  0.975\tiny$\pm$0.000 &
  \best{2.258}\tiny$\pm$0.000 &
  \best{0.930}\tiny$\pm$0.000 &
  2.467\tiny$\pm$0.000 &
  1.079\tiny$\pm$0.000 &
  2.288\tiny$\pm$0.000 &
  0.962\tiny$\pm$0.000 \\
 &
  36 &
  \best{2.162}\tiny$\pm$0.000 &
  \best{0.940}\tiny$\pm$0.000 &
  2.292\tiny$\pm$0.000 &
  0.986\tiny$\pm$0.000 &
  \second{2.213}\tiny$\pm$0.000 &
  \second{0.943}\tiny$\pm$0.000 &
  2.510\tiny$\pm$0.000 &
  1.089\tiny$\pm$0.000 &
  2.221\tiny$\pm$0.000 &
  0.964\tiny$\pm$0.000 \\
 &
  48 &
  \second{2.184}\tiny$\pm$0.000 &
  \best{0.956}\tiny$\pm$0.000 &
  2.266\tiny$\pm$0.000 &
  0.992\tiny$\pm$0.000 &
  2.309\tiny$\pm$0.000 &
  0.979\tiny$\pm$0.000 &
  2.510\tiny$\pm$0.000 &
  1.092\tiny$\pm$0.000 &
  \best{2.177}\tiny$\pm$0.000 &
  \second{0.970}\tiny$\pm$0.000 \\
 &
  60 &
  \best{2.053}\tiny$\pm$0.000 &
  \second{0.957}\tiny$\pm$0.000 &
  2.221\tiny$\pm$0.000 &
  0.996\tiny$\pm$0.000 &
  \second{2.183}\tiny$\pm$0.000 &
  \best{0.952}\tiny$\pm$0.000 &
  2.599\tiny$\pm$0.000 &
  1.121\tiny$\pm$0.000 &
  2.216\tiny$\pm$0.000 &
  0.985\tiny$\pm$0.000 \\
\midrule
\multirow{4}{*}{M5} &
  24 &
  \best{0.490}\tiny$\pm$0.000 &
  \best{0.497}\tiny$\pm$0.000 &
  0.499\tiny$\pm$0.000 &
  \second{0.502}\tiny$\pm$0.000 &
  0.500\tiny$\pm$0.000 &
  0.503\tiny$\pm$0.000 &
  0.503\tiny$\pm$0.000 &
  0.505\tiny$\pm$0.000 &
  \second{0.498}\tiny$\pm$0.000 &
  \second{0.502}\tiny$\pm$0.000 \\
 &
  36 &
  \best{0.515}\tiny$\pm$0.000 &
  \best{0.511}\tiny$\pm$0.000 &
  0.523\tiny$\pm$0.000 &
  0.529\tiny$\pm$0.000 &
  0.524\tiny$\pm$0.000 &
  0.517\tiny$\pm$0.000 &
  0.533\tiny$\pm$0.000 &
  0.521\tiny$\pm$0.000 &
  \second{0.522}\tiny$\pm$0.000 &
  \second{0.515}\tiny$\pm$0.000 \\
 &
  48 &
  \best{0.542}\tiny$\pm$0.000 &
  \best{0.525}\tiny$\pm$0.000 &
  0.550\tiny$\pm$0.000 &
  0.530\tiny$\pm$0.000 &
  0.552\tiny$\pm$0.000 &
  0.531\tiny$\pm$0.000 &
  0.563\tiny$\pm$0.000 &
  0.537\tiny$\pm$0.000 &
  \second{0.549}\tiny$\pm$0.000 &
  \second{0.529}\tiny$\pm$0.000 \\
 &
  60 &
  \best{0.565}\tiny$\pm$0.000 &
  \best{0.538}\tiny$\pm$0.000 &
  0.572\tiny$\pm$0.000 &
  0.542\tiny$\pm$0.000 &
  0.575\tiny$\pm$0.000 &
  0.543\tiny$\pm$0.000 &
  0.581\tiny$\pm$0.000 &
  0.547\tiny$\pm$0.000 &
  \second{0.570}\tiny$\pm$0.000 &
  \second{0.541}\tiny$\pm$0.000 \\
\midrule
\multicolumn{2}{c}{Best count} &
  \best{32} &
  \best{26} &
  \second{7} &
  5 &
  \second{7} &
  \second{7} &
  6 &
  1 &
  9 &
  6 \\
\bottomrule
\end{tabular}
}
\end{table}

We first compared our model \ours\ with other leading FC models. To ensure a fair evaluation, all models were trained to converge under consistent training settings and using optimal hyperparameters from the authors’ original implementations. Table \ref{tab:2} presents the evaluation results for FC models on the LTSF and STSF datasets.

Across all eight datasets, \ours\ consistently delivers comparable or even superior performance. On datasets with smaller sample sizes (ETTh1, ETTh2, Faas, Iaas, Illness, and M5), our model significantly outperforms its counterparts, demonstrating its ability to effectively capture temporal patterns with limited training data while avoiding overfitting. On larger datasets (ETTm1, ETTm2, Electricity, and Weather), \ours\ achieves results comparable to other linear models. 

One notable advantage of \ours\ is its substantially reduced parameter count compared to FC models. By operating within a low-dimensional manifold in the FC models' model space, \ours\ achieves performance comparable to or better than FC models, suggesting that our carefully designed model space provides a highly efficient and effective solution for these LTSF benchmark tasks. This highlights the robustness and effectiveness of our architecture in representing the essential solution space of FC models with minimal complexity.

\subsubsection{Versus nonlinear models.}

\begin{table}[htbp]
\centering
\caption{Results compared with non linear models. The best and second results are highlighted in \best{bold} and \second{underline}.}
\resizebox{\textwidth}{!}{
\begin{tabular}{cccccccccccccccccccccc}
\toprule
 &
   &
  \multicolumn{2}{c}{\textbf{\ours}} &
  \multicolumn{2}{c}{TimeMixer\texttt{++}} &
  \multicolumn{2}{c}{TQNet} &
  \multicolumn{2}{c}{iTransformer}&
  \multicolumn{2}{c}{PatchTST} &
  \multicolumn{2}{c}{TimesNet} &
  \multicolumn{2}{c}{Crossformer} &
  \multicolumn{2}{c}{MICN} &
  \multicolumn{2}{c}{FEDformer} \\
 &
   &
  \multicolumn{2}{c}{\textbf{(Ours)}} &
  \multicolumn{2}{c}{(2025)} &
  \multicolumn{2}{c}{(2025)} &
  \multicolumn{2}{c}{(2024)} &
  \multicolumn{2}{c}{(2023)} &
  \multicolumn{2}{c}{(2023)} &
  \multicolumn{2}{c}{(2023)} &
  \multicolumn{2}{c}{(2023)} &
  \multicolumn{2}{c}{(2022)} \\
\midrule
\multicolumn{2}{c}{Metrics $\downarrow$} & MSE            & MAE     & MSE            & MAE           & MSE            & MAE            & MSE            & MAE            & MSE            & MAE            & MSE   & MAE   & MSE   & MAE  & MSE   & MAE  & MSE   & MAE\\
\midrule
\multirow{4}{*}{ETTh1} 
    & 96  
        & \second{0.369} & \best{0.393} 
        &\best{0.361}&0.403
        & 0.371 & \best{0.393}
        & 0.386 & 0.405          
        & 0.370 & \second{0.399}
        & 0.384 & 0.402 
        & 0.386 & 0.429 
        & 0.396 & 0.427 
        & 0.376 & 0.415\\
    & 192 
        & \best{0.407} & \best{0.415} 
        &0.416&0.441
        &0.428 & 0.426
        & 0.424          & 0.440          
        & \second{0.413}    & \second{0.421}    
        & 0.557 & 0.436 
        & 0.419 & 0.444 
        & 0.430 & 0.453 
        & 0.423 & 0.446\\
    & 336 
        & \second{0.424}    & \best{0.427} 
        &0.430&0.434
        &0.476 & 0.446
        & 0.449          & 0.460          
        & \best{0.422} & \second{0.436}    
        & 0.491 & 0.469 
        & 0.440 & 0.461 
        & 0.433 & 0.458 
        & 0.444 & 0.462\\
    & 720 
        & \best{0.409} & \best{0.439} 
        &0.467&\second{0.451}
        &0.487 &0.470
        & 0.495          & 0.487          
        & \second{0.447}    & 0.466   
        & 0.521 & 0.500 
        & 0.519 & 0.524 
        & 0.474 & 0.508 
        & 0.469 & 0.492\\
\midrule
\multirow{4}{*}{ETTh2} 
    &96 
      &\second{0.275} &  \second{0.336 }
      & 0.276 & \best{0.328}
      & 0.295 & 0.343
      &      0.297 &  0.348 
      &      \best{0.274} &  \second{0.336} 
      &      0.340 &  0.374 
      &      0.276 &  0.338
      & 0.289& 0.357 
      & 0.332 & 0.374\\
    & 192 
        & \best{0.325} & \best{0.372} 
        & 0.342 & \second{0.379} 
        &0.367 &0.393
        & 0.371 & 0.403          
        & \second{0.339}    & \second{0.379}    
        & 0.402 & 0.414 
        & 0.345 & 0.382 
        & 0.409& 0.438
        & 0.407 & 0.446\\
    & 336 
        & 0.350    & \second{0.393}    
        & \second{0.346}             & 0.398
        &0.417              &0.427
        & 0.404          & 0.428          
        & \best{0.329} & \best{0.380} 
        & 0.452 & 0.452 
        & 0.375 & 0.411 
        & 0.417 & 0.452 
        & 0.400 & 0.447\\
    & 720 
        & \best{0.375} & \best{0.415} 
        & 0.392 & 0.415
        &0.433&0.446
        & 0.424          & 0.444          
        & \second{0.379}    & \second{0.422}    
        & 0.462 & 0.468 
        & 0.408 & 0.446 
        & 0.426 & 0.473 
        & 0.412 & 0.469\\
\midrule
\multirow{4}{*}{ETTm1} 
    & 96  
        & 0.309          & 0.350   
        &0.310 &\best{0.334} 
        &0.311&0.353
        & \second{0.300}    & 0.353          
        & \best{0.290} & \second{0.342} 
        & 0.338 & 0.375 
        & 0.316 & 0.373 
        & 0.314 & 0.360 
        & 0.326 & 0.390\\
                       
    & 192 
        & \second{0.339}    & \second{0.369} 
        & 0.348 & \best{0.362}
        &0.356&0.378
        & 0.345          & 0.382   
        & \best{0.332} & \second{0.369} 
        & 0.371 & 0.387 
        & 0.377 & 0.411 
        & 0.359 & 0.387 
        & 0.365 & 0.415\\
        
    & 336 
        & \best{0.367} & \best{0.386} 
        &0.376 & \second{0.391}
        &0.390&0.401
        & 0.374          & 0.398          
        & \second{0.369}    & 0.392    
        & 0.410 & 0.411 
        & 0.431 & 0.442 
        & 0.398 & 0.413 
        & 0.392 & 0.425\\
                       
    & 720 
        & \best{0.416} & \best{0.413} 
        &0.440 & 0.423
        &0.452&0.440
        & \second{0.429}    & 0.430    
        & \best{0.416} & \second{0.420}    
        & 0.478 & 0.450 
        & 0.600 & 0.547 
        & 0.459 & 0.464 
        & 0.446 & 0.458\\
        
\midrule

\multirow{4}{*}{ETTm2} 
    & 96  
        & \best{0.162} & \second{0.252 }
        &0.170 &\best{0.245}
        &0.173&0.256
        & 0.175          & 0.266          
        & \second{0.165}    & 0.255    
        & 0.187 & 0.267 
        & 0.421 & 0.461 
        & 0.178 & 0.273 
        & 0.180 & 0.271\\
                       
    & 192 
        & \best{0.216} & \best{0.289} 
        &0.229 & \second{0.291} 
        &0.238&0.298
        & 0.242          & 0.312          
        & \second{0.220}    & 0.292   
        & 0.249 & 0.309 
        & 0.503 & 0.519 
        & 0.245 & 0.316 
        & 0.252 & 0.318\\
                       
    & 336 
        & \best{0.268} & \best{0.324} 
        &0.303 &0.343
        &0.301&0.340
        & 0.282          & 0.340          
        & \second{0.274}    & \second{0.329}    
        & 0.321 & 0.351 
        & 0.611 & 0.580  
        & 0.295 & 0.350 
        & 0.324 & 0.364\\

    & 720 
        & \best{0.353} & \best{0.379}  
        &0.373 & 0.399
        &0.397&0.396
        & 0.378          & 0.398          
        & \second{0.362}    & \second{0.385}    
        & 0.497 & 0.403 
        & 0.996 & 0.750  
        & 0.389 & 0.406 
        & 0.410 & 0.420\\
        
\midrule

\multirow{4}{*}{Electricity} 

    &96 
        &  \second{0.132} &  \second{0.228} 
        &0.135 &\best{0.222}
        &0.134  &0.229
        &  \second{0.132} &  0.229 
        &  \best{0.129} &  \best{0.222 }
        &0.168 &  0.272 
        &0.187 &  0.283 
        & 0.159 & 0.267 
        & 0.186 & 0.302\\
        
    & 192 
        & \second{0.148}    & 0.245  
        &\best{0.147} & \best{0.235}
        &0.154&0.247
        & 0.151          & 0.246          
        & \best{0.147} & \second{0.240 }  
        & 0.184 & 0.289 
        & 0.258 & 0.330 
        & 0.168 & 0.279 
        & 0.197 & 0.311\\
        
    & 336 
        & \best{0.162} & 0.261   
        &0.164 &\best{0.245}
        &0.169&0.264
        & 0.167          & 0.264          
        & \second{0.163}    & \second{0.259 } 
        & 0.198 & 0.300 
        & 0.323 & 0.369 
        & 0.196 & 0.308 
        & 0.213 & 0.328\\
        
    & 720 
        & 0.198          & 0.296       
        &0.212 &0.310
        &0.201&0.294
        & \best{0.194} & \best{0.286} 
        & \second{0.197}    & \second{0.290}    
        & 0.220 & 0.320 
        & 0.404 & 0.423
        & 0.203 & 0.312
        & 0.233 & 0.344\\
        
\midrule

\multirow{4}{*}{Weather} 

    &96 
        &  \best{0.142} &  0.201 
        &0.155 &0.205
        &0.157&\second{0.200}
        &  0.159 &  0.208 
        &\second{0.149} &  \best{0.198} 
        &0.172 &  0.220 
        &  0.153 &  0.217 
        & 0.161 & 0.226 
        & 0.238 & 0.314\\
        
    & 192 
        & \best{0.187} & 0.253 
        &0.201 &\second{0.245}
        &0.206&\second{0.245}
        & 0.200          & 0.248 
        & \second{0.194}    & \best{0.241}
        & 0.219 & 0.261 
        & 0.197 & 0.269
        & 0.220 & 0.283
        & 0.275 & 0.329\\
        
    & 336 
        & \best{0.234} & 0.293  
        &0.237 &\best{0.265}
        &0.262&0.287
        & 0.253          & 0.289
        & \second{0.245}    & \second{0.282}
        & 0.280 & 0.306 
        & 0.252 & 0.311 
        & 0.275 & 0.328 
        & 0.339 & 0.377\\
        
    & 720 
        & \best{0.306} & 0.348    
        & 0.312 & \best{0.334}
        &0.344&0.342
        & 0.321          & \second{0.338} 
        & 0.314    & \best{0.334} 
        & 0.365 & 0.359 
        & 0.318 & 0.363 
        & \second{0.311} & 0.356 
        & 0.389 & 0.409\\
        
\midrule

\multicolumn{2}{c}{Best count} 

    &  \best{15} &  \best{11} 
    &2 &\second{9}
    &0 &1
    &  1 &  1 
    &  \second{8} &  5 
    &  0 &  0 
    &  0&  0 
    & 0 & 0 
    & 0 & 0\\
    
\bottomrule
\end{tabular}
}
\label{tab:3}
\end{table}

Next, we compare \ours\ with several nonlinear models. Table \ref{tab:3} presents the evaluation results for these nonlinear models specifically on the public LTSF datasets. We do not report model performance on the private LTSF and STSF datasets due to the following practical limitations: for private LTSF datasets, excessively long input lengths lead to high memory usage for nonlinear models, resulting in prohibitively high deployment costs in production environments. For STSF datasets, the limited amount of training data makes these models highly prone to overfitting, which complicates the training process.

To ensure a rigorous comparison, baseline models were evaluated by selecting the best result from a comprehensive search over input lengths $L\in \{96,192,336,512,672,720\}$. To benchmark against the very latest advancements, we also include two recent models, TimeMixer\texttt{++}\cite{wang2025timemixer} and TQNet\cite{lin2025TQNet}, citing the best performance reported in their original publications.

The results in Table \ref{tab:3} highlight the remarkable effectiveness of \ours. Notably, our model establishes a new state-of-the-art, achieving the best overall accuracy against all competing methods. It significantly outperforms established baselines like iTransformer and recent models such as TimeMixer\texttt{++}\cite{wang2025timemixer} and TQNet\cite{lin2025TQNet}. This SOTA performance is achieved with unparalleled efficiency. For instance, \ours\ matches or exceeds the accuracy of PatchTST \cite{dosovitskiy2021an} while utilizing only $10^{-4}$ of its parameters. This massive reduction in model complexity, combined with its superior forecasting accuracy, underscores the competitive advantage of \ours\ and establishes it as a new, highly efficient benchmark for LTSF.

Furthermore, our state-of-the-art results, achieved with a purely linear model, contribute strong evidence to the ongoing debate \cite{zeng2023transformers} regarding the necessity of complex, non-linear architectures for LTSF. The success of \ours~suggests that, for these benchmarks, a well-structured {inductive bias} (i.e., the factorization of frequential and temporal components) is {more critical for generalization} than the capacity for high-dimensional non-linear mapping (as found in Transformers).

\subsubsection{Parameter complexity and efficiency.}

\begin{figure*}[htbp]
    \centering

    \begin{minipage}{\textwidth}
        \centering
        \includegraphics[width=0.7\textwidth]{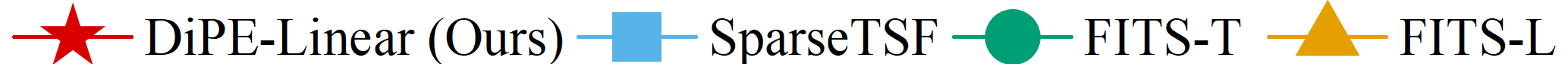}
    \end{minipage}

    \vfill %

    \subfloat[ETTh1\label{fig:5a}]{%
        \includegraphics[width=0.24\textwidth]{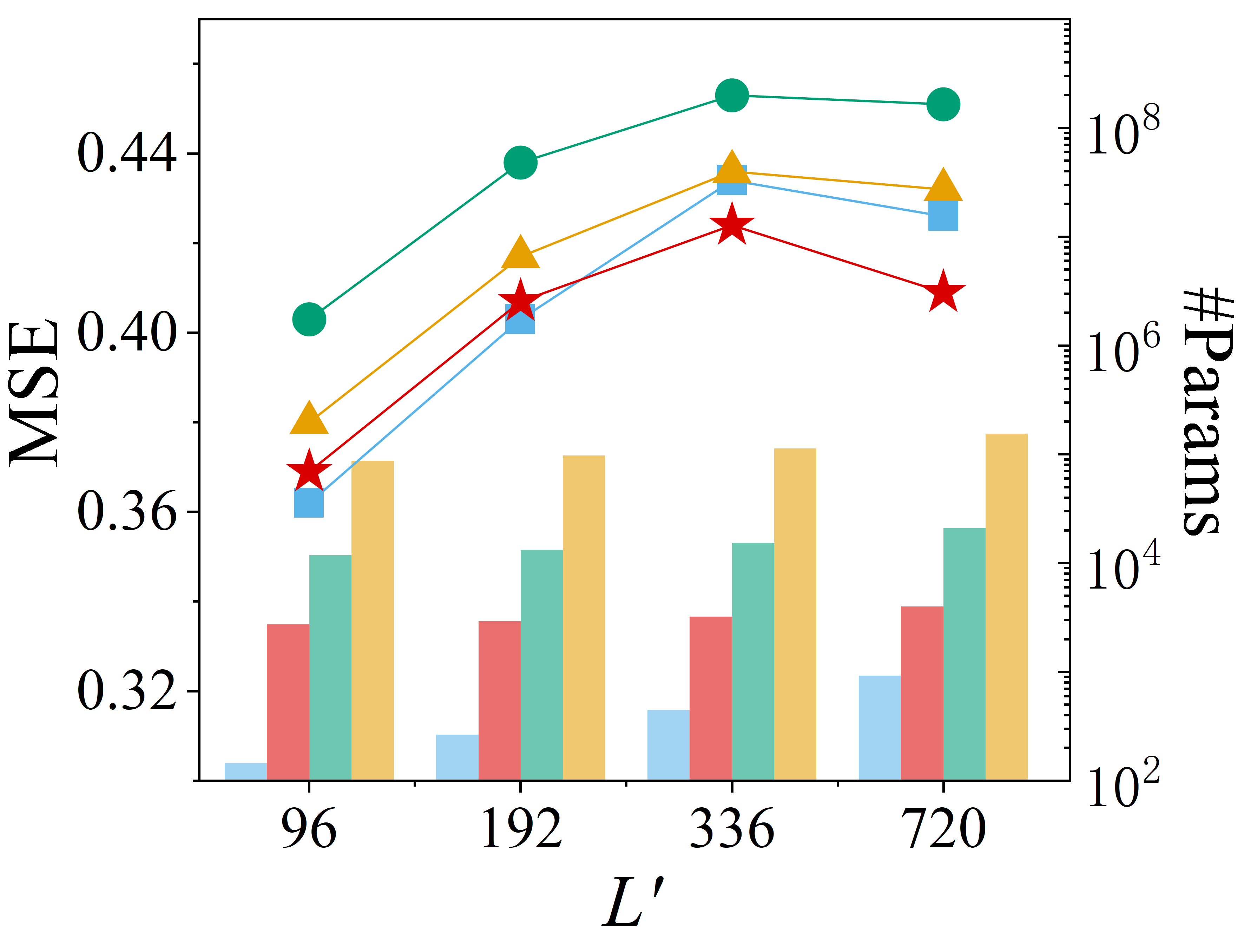}%
    }
    \hfill %
    \subfloat[ETTm1\label{fig:5b}]{%
        \includegraphics[width=0.24\textwidth]{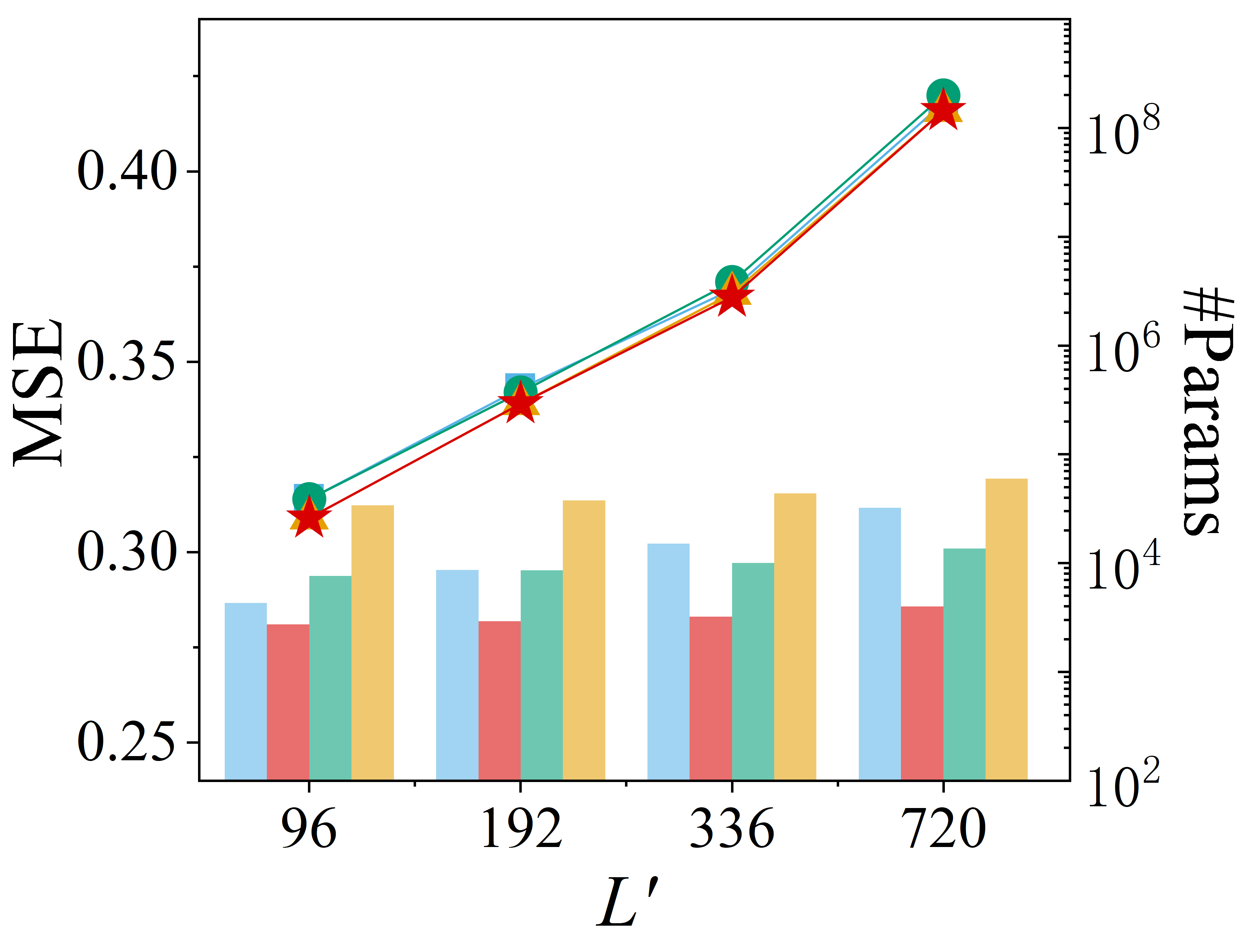}%
    }
    \hfill %
    \subfloat[Electricity\label{fig:5c}]{%
        \includegraphics[width=0.24\textwidth]{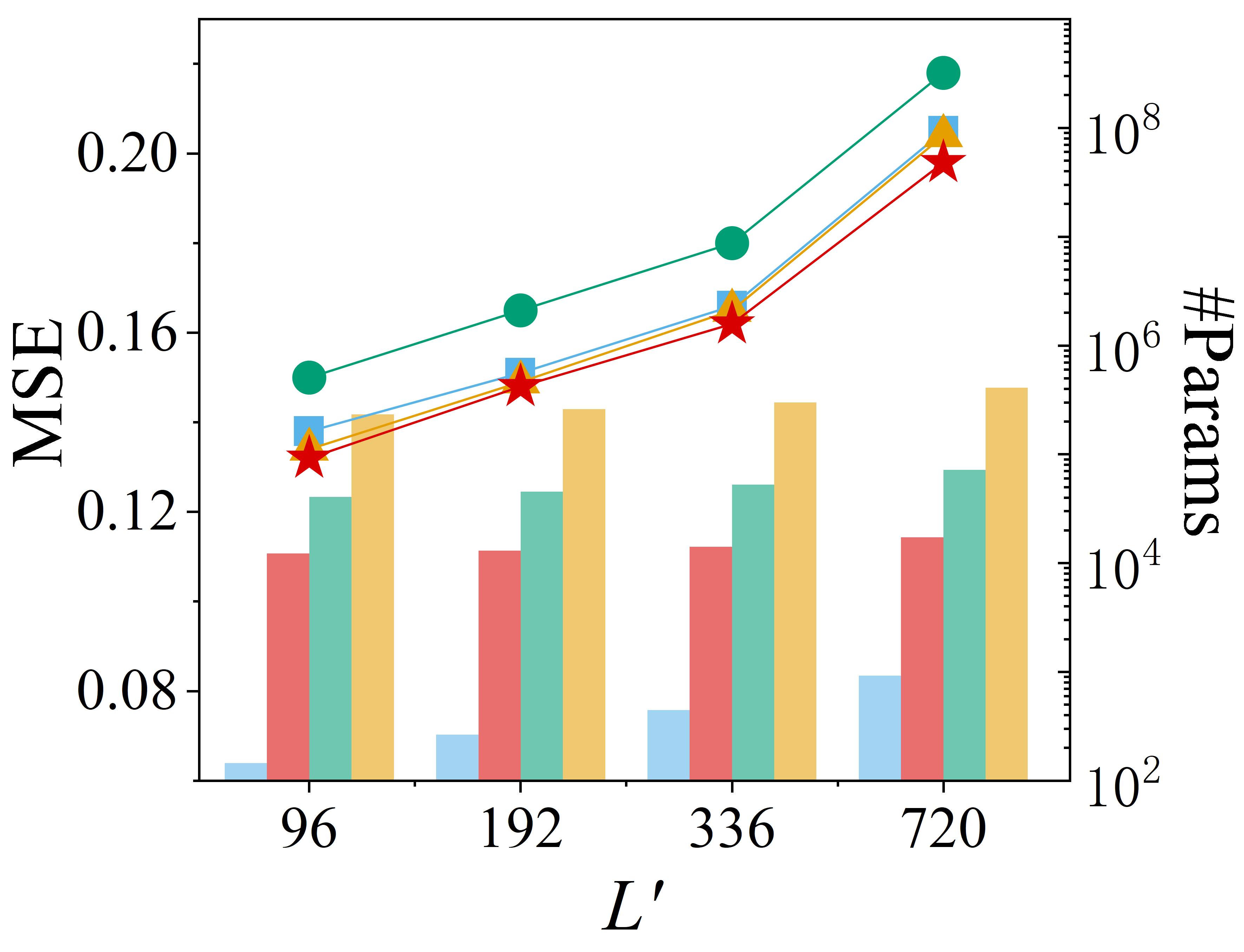}%
    }
    \hfill %
    \subfloat[Weather\label{fig:5d}]{%
        \includegraphics[width=0.25\textwidth]{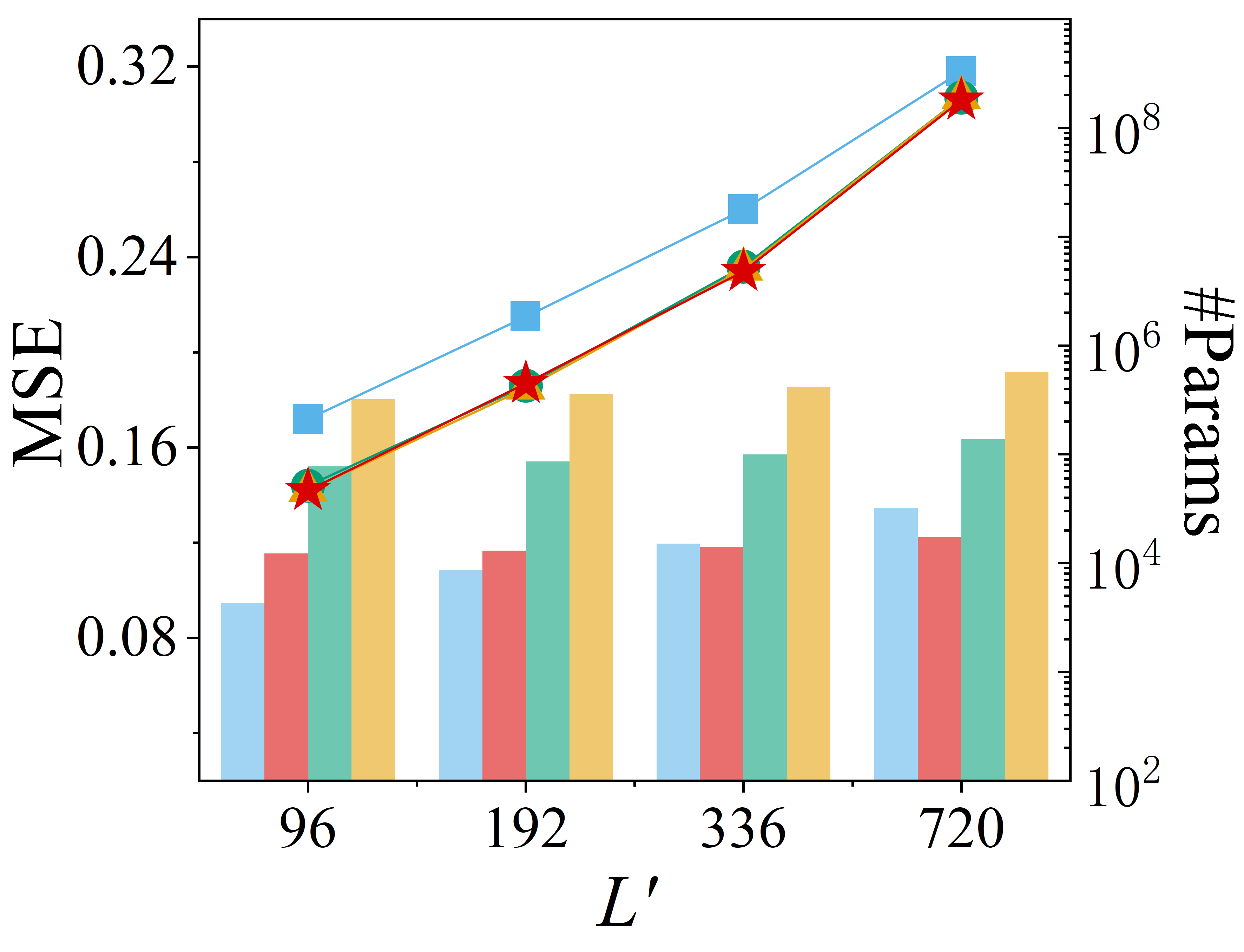}%
    }

    \caption{Comparison of the number of parameters (bars) and MSE (lines) between parameter-efficient models.}
    \label{fig:sparse}
\end{figure*}

Recent studies have focused on parameter efficient models. We compare \ours\ with SparseTSF \cite{pmlr-v235-lin24n} and FITS \cite{xu2023fits} (FITS-T and FITS-L, representing different low-pass filter settings). As shown in Fig.~\ref{fig:sparse}, our model achieves the lowest MSE across these parameter-efficient baselines. Notably, \ours\ not only uses fewer parameters than FITS-T but also matches or outperforms FITS-L, demonstrating a superior balance of performance and parameter efficiency.

Fig.~\ref{fig:mse-param} further demonstrates the effectiveness of \ours\ by comparing its average MSE and parameter counts with other baseline models across all prediction lengths and datasets. The results highlight that \ours\ achieves state-of-the-art performance while maintaining a drastically smaller parameter count, underscoring its efficiency and practicality.

\subsubsection{Effect of Training Data Size.}
\begin{figure}[htbp]
    \centering

    \subfloat[ETTh1\label{fig:7a}]{%
        \includegraphics[width=0.4\columnwidth]{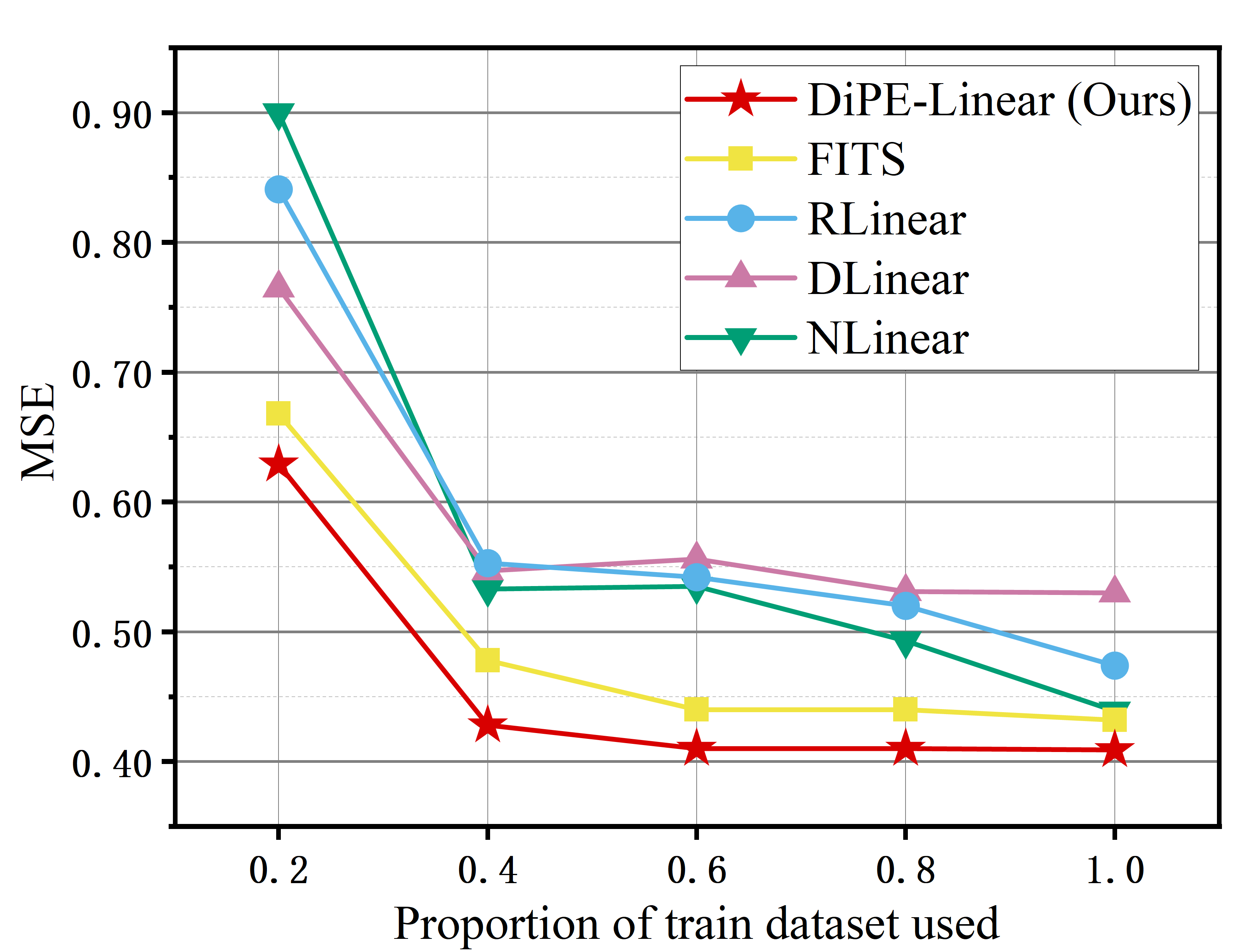}%
    }
    \hspace{0.1\linewidth}
    \subfloat[ETTm1\label{fig:7b}]{%
        \includegraphics[width=0.4\columnwidth]{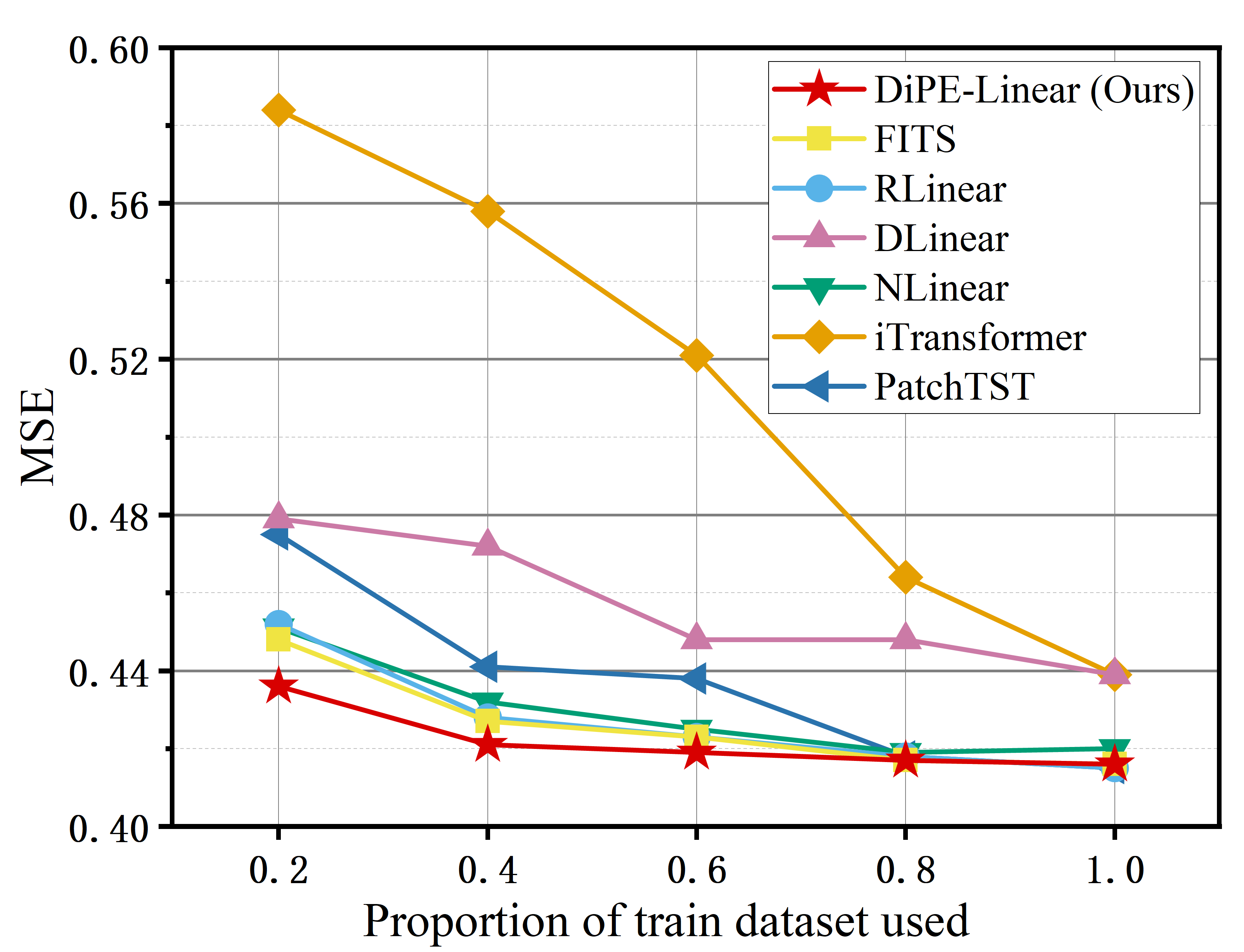}%
    }

    \caption{MSE with varying proportion of ETT train datasets.}
    \label{fig:proportion}
\end{figure}

To further investigate whether the model can perform effectively with reduced training data, we conducted experiments on the ETTh1 and ETTm1 datasets using different proportions of the training set. For the ETTh1 dataset, we compared our model with other FC models. Additionally, since the ETTm1 dataset contains more samples, we included nonlinear models for a broader comparison. As shown in Fig.~\ref{fig:proportion}, \ours\ outperforms both FC models and nonlinear models across all proportions of the training set. Notably, on the ETTh1 dataset, our model surpasses the performance of other FC models trained on the full dataset while using only 40\% of the training data.

\subsection{Ablation Studies}

\subsubsection{Basic Components.}
To comprehensively validate the effectiveness of our model, we conducted ablation tests on two submodules: SFA and STA. The submodule IFM, which maps the input sequence space to the output sequence space, serves as the core component of our model and therefore cannot be ablated. Table \ref{tab:5} presents the results of the ablation experiments on the ETTh2 and Weather datasets with various prediction lengths. In both the ETTh2 and Weather datasets, the incorporation of the SFA and STA modules generally enhances the model's prediction accuracy. Specifically, in the Weather dataset, the abundance of training samples ensures that each submodule effectively captures significant patterns, leading to a substantial improvement in prediction performance.  This demonstrates the necessity of addressing the importance of both the frequency domain and time domain in our input data processing.

\begin{table}[htp]
\centering
\caption{Ablation on components.  MSE is used as the evaluation metric, and the best result in each row is highlighted in \textbf{bold}.}
\label{tab:5}
\begin{tabular}[tb]{ccccccccccc}
\toprule
\multicolumn{3}{c}{Components}&\multicolumn{4}{c}{ETTh2}&\multicolumn{4}{c}{Weather}\\
\midrule
SFA&STA&IFM& 96 & 192 & 336& 720 &96 & 192 & 336& 720 \\
\midrule

 &  & \cmark &0.295&0.340&0.352&0.387   &0.175&0.228&0.274&0.332\\

 & \cmark & \cmark &0.284&0.331&\textbf{0.348}&0.382   &0.146&0.191&0.243&0.315\\
\cmark &  & \cmark &0.290&0.331&\textbf{0.348}&0.379   &0.166&0.220&0.272&0.331\\
\midrule
\cmark & \cmark & \cmark &\textbf{0.275}&\textbf{0.325}&0.350&\textbf{0.375}   &\textbf{0.142}&\textbf{0.187}&\textbf{0.234}&\textbf{0.306}\\
\bottomrule
\end{tabular}
\end{table}

\subsubsection{Hyperparameter Analysis and Guidelines.}

We analyze the sensitivity of the two key hyperparameters, $M$ and $\alpha$, to provide practical tuning guidelines and address concerns about model complexity.

\begin{figure}[htbp]
    \centering

    \subfloat[Weather\label{fig:4x}]{%
        \includegraphics[width=0.33\textwidth]{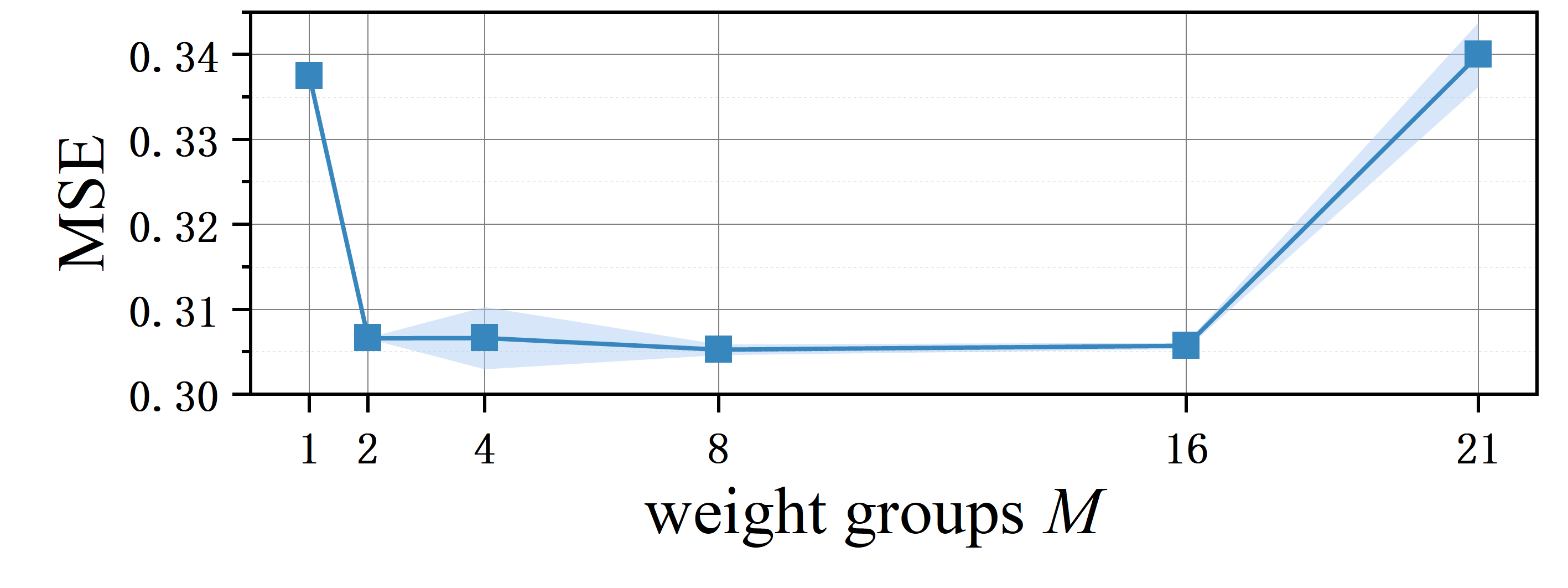}%
    }
    \hfill
    \subfloat[ETTh1\label{fig:4a}]{%
        \includegraphics[width=0.33\textwidth]{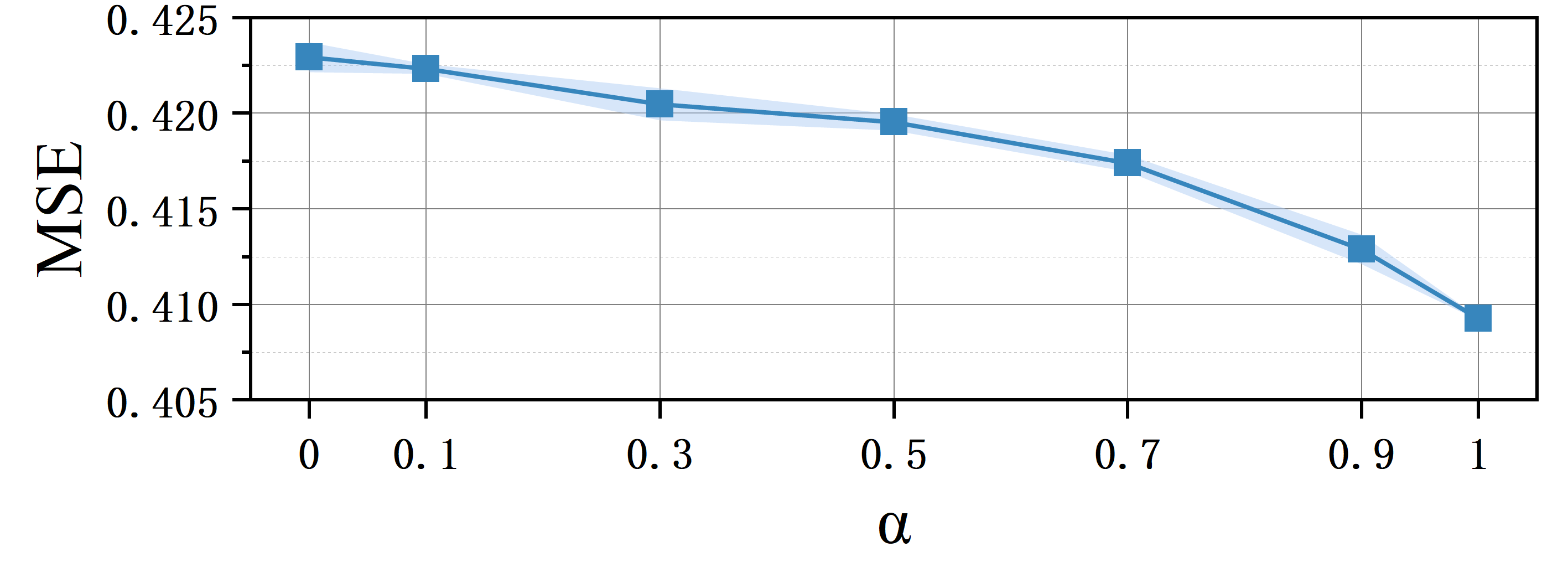}%
    }
    \hfill
    \subfloat[ETTh2\label{fig:4b}]{%
        \includegraphics[width=0.33\textwidth]{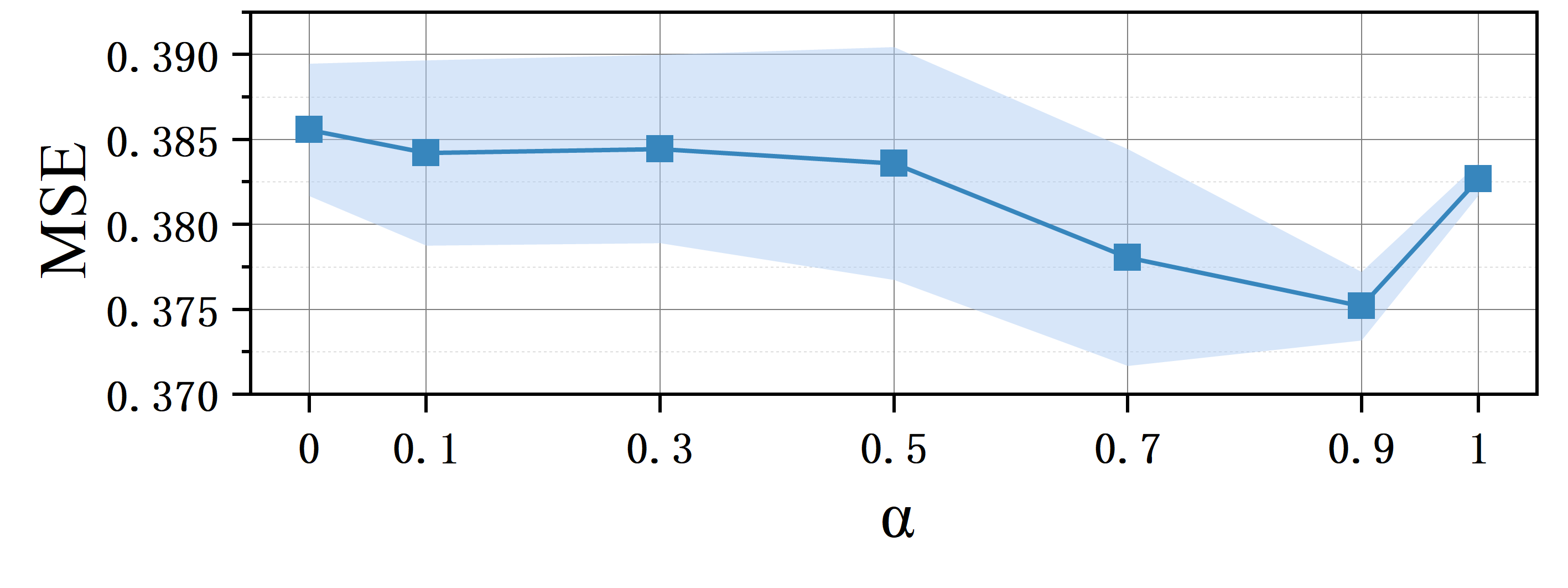}%
    }

    \caption{Ablation MSE on different $\alpha$ and $M$ setting on LTSF datasets. The forecasting horizon $L^\prime=720$. Mean and standard deviation are calculated over five independent runs.}
    \label{fig:6}
\end{figure}

For the {Low-rank weight sharing factor} $M$, this hyperparameter can be intuitively understood as the number of distinct patterns among the variables in the multivariate time series. Fig.~\ref{fig:4x} presents the results of varying $M$ on the Weather dataset ($L^\prime=720$). The key observation is that model performance remains stable and near-optimal across a wide range of $M$ (from 2 to 16). Performance only degrades at the extremes: $M=1$ (full weight sharing) and $M=21$ (full channel independence). This demonstrates that $M$ {is not a sensitive hyperparameter} within a reasonable range, significantly easing the tuning process. In practice, we suggest starting $M=1$ and increasing it exponentially(as $M=1, 2, 4, \dots$) until validation performance saturates.

For the {SFALoss balance parameter} $\alpha$, which governs the trade-off between the time-domain loss ($\alpha=0$) and the frequency-domain loss ($\alpha=1$), we performed a grid search on the ETTh datasets (results in Fig.~ \ref{fig:6}). Our experiments show that the model's accuracy often has a monotonic or U-shaped relationship with $\alpha$. Notably, the optimal performance is typically found when $\alpha$ is in the ranges of {[0, 0.3] or [0.7, 1.0]}. This suggests the model prefers to optimize primarily for one domain (either time or frequency) rather than an equal mixture. This finding simplifies tuning, as the search space can be effectively concentrated on values near 0 and 1, rather than requiring a fine-grained search across the entire [0, 1] interval.

\subsection{Complexity Analysis}

Our model retains the linear parameterization of FC models while significantly reducing both parameter and computational complexity. The parameter complexity in the SFA, STA, and IFM modules is \(\mathcal O(M \cdot L)\). Low-rank weight sharing introduces additional routing parameters with a complexity of \(\mathcal O(M \cdot C)\), where \(C\) is the number of channels; this is a higher-order infinitesimal relative to \(\mathcal O(M \cdot L)\). Thus, the overall parameter complexity remains \(\mathcal O(M \cdot L)\). For computational complexity, the model's main costs arise from element-wise multiplications (\(\mathcal O(C \cdot L)\)) and the rFFT/irFFT operations (\(\mathcal O(C \cdot L \log L)\)), resulting in a total computational complexity of \(\mathcal O(C \cdot L \log L)\). Table \ref{tab:complexity} compares these complexities with those of standard FC models, demonstrating the efficiency of our approach.

We note that while our theoretical computational complexity is superior, practical latency depends on hardware optimizations; FC models benefit from highly optimized matrix multiplication (e.g. cuBLAS), which may not be true for FFT operations. Therefore, the primary advantages of \ours\ are its {linear parameter complexity} ($\mathcal{O}(M \cdot L)$) and {drastically reduced memory footprint}. This makes \ours\ ideal for resource-constrained environments and for mitigating overfitting on long sequences.

\begin{table}[htbp]
\centering
\caption{Comparison of parameter and time complexity with FC models. Note that $M \ll C \ll L$ in LTSF settings.}
\label{tab:complexity}
\begin{tabular}{cccc}
\toprule
        & \textbf{\ours} (Ours)       & FC (w/ sharing)                 & FC (w/o sharing)                 \\
\midrule
Parameter  & $\mathcal O(M\cdot L)$    & $\mathcal O(L^2)$    & $\mathcal O(C\cdot L^2)$ \\
Time & $\mathcal O(C\cdot L\log L)$ & $\mathcal O(C\cdot L^2)$ & $\mathcal O(C\cdot L^2)$\\
\bottomrule
\end{tabular}
\end{table}

\section{Conclusion}

This paper introduces \ours, a novel model specifically tailored for LTSF. \ours\ addresses the intrinsic quadratic parameter redundancy and {entanglement} of existing {monolithic} FC models by employing disentangled modules---factorizing the modeling task into specialized components for frequential filtering, temporal weighting, and frequential mapping. This innovative architecture, which serves as a strong {inductive bias}, reduces parameter complexity from quadratic to linear and computational complexity from quadratic to log-linear, significantly improving efficiency and scalability. Comprehensive experimental evaluations demonstrate that \ours\ consistently achieves or surpasses state-of-the-art models, suggesting that for LTSF benchmarks, its well-structured linear design is more effective for generalization than the complex non-linear mapping of larger models. \ours~thus sets a new baseline for parameter-efficient LTSF models. Future research will explore the integration of nonlinear architectures and further optimization of efficient LTSF frameworks.

\bibliographystyle{splncs04}
\bibliography{ijcai25}

\end{document}